\documentclass{article}

\usepackage{arxiv}
\usepackage{microtype}
\usepackage{graphicx}
\usepackage{multirow}
\usepackage{booktabs} 
\usepackage{subcaption}
\usepackage{algorithm}
\usepackage{algorithmic}
\usepackage[numbers]{natbib}
\usepackage{hyperref}

\newcommand{\nodeact}{node\_action tuple}
\newcommand{\nodeacts}{node\_action tuples}

\newcommand{\searchalg}{Q*}
\newcommand{\bwsearchalg}{BWQS}

\usepackage{amsmath,amsfonts,amsthm,mathabx,amssymb}

\DeclareMathOperator{\ctgfunc}{\mathit{j_\theta}}
\DeclareMathOperator{\ctgparams}{\mathit{\theta}}
\DeclareMathOperator{\dqnfunc}{\mathit{q_\phi}}
\DeclareMathOperator{\dqnparams}{\mathit{\phi}}

\usepackage[english]{babel}
\usepackage{amsthm}
\newtheorem{theorem}{Theorem}
\theoremstyle{case}

\newtheorem{lemmaT}{Lemma}

\usepackage{xcolor}
\usepackage{tikz}
\usepackage{subcaption}
\usepackage{adjustbox}
\usetikzlibrary{positioning}
\tikzset{
	layer/.style={draw, minimum width=4cm, minimum height=1cm, align=center},
	resblock/.style={draw, dashed, minimum width=4cm, minimum height=1cm, align=center},
	arrow/.style={->, thick}
}

\usepackage{siunitx}

\newcommand{\nk}[1]{\num[round-mode=places,round-precision=0]{#1}k}

\usepackage{xspace}
\newcommand{\start} {\mbox{$\mathit{start}$}\xspace} 
\newcommand{\goal} {\mbox{$\mathit{goal}$}\xspace} 

\usepackage{multirow}
\usepackage{subcaption}
\usepackage{booktabs}       

\usepackage{longtable}

\title{A* Search Without Expansions: Learning Heuristic Functions with Deep Q-Networks}

\author{
	Forest Agostinelli\\
	Department of Computer Science and Engineering\\
	University of South Carolina\\
	\texttt{foresta@cse.sc.edu} \\
	\And
	Shahaf S. Shperberg\\
	Faculty of Computer and Information Science\\
	Ben-Gurion University of the Negev\\
	\texttt{shperbsh@bgu.ac.il} \\
	\And
	Alexander Shmakov \\
	Department of Computer Science\\
	University of California, Irvine\\
	\texttt{ashmakov@uci.edu} \\
	\And
	Stephen McAleer\\
	Department of Computer Science\\
	University of California, Irvine\\
	\texttt{smcaleer@uci.edu} \\
	\And
	Roy Fox\\
	Department of Computer Science\\
	University of California, Irvine\\
	\texttt{royf@uci.edu} \\
	\And
	Pierre Baldi\\
	Department of Computer Science\\
	University of California, Irvine\\
	\texttt{pfbaldi@uci.edu}\\
}
\date{}

\begin{document}

\maketitle

\begin{abstract}
Efficiently solving problems with large action spaces using A* search remains a significant challenge. This is because, for each iteration of A* search, the number of nodes generated and the number of heuristic function applications grow linearly with the size of the action space. This burden becomes even more apparent when A* search uses a heuristic function learned by computationally expensive function approximators, such as deep neural networks. To address this issue, we introduce \searchalg{}, a search algorithm that leverages heuristics capable of receiving a state and, in a single function call, returning cost-to-go estimates for all possible transitions from that state, along with estimates of the corresponding transition costs---without the need to apply the transitions or generate the successor states; such action-state estimation are typically known as Q-values. This significantly reduces computation time and memory usage. In addition, we prove that \searchalg{} search is guaranteed to find a shortest path given a heuristic function that does not overestimate the sum of the transition cost and cost-to-go of the state.
To obtain heuristics for Q* search, we employ a deep Q-network architecture to learn a state-action heuristic function from domain interaction, without any prior knowledge.
We use \searchalg{} with our learned heuristic on different domains and action spaces, showing that \searchalg{} suffers from only a small runtime overhead as the size of the action space increases. In addition, our empirical results show \searchalg{} search is up to 129 times faster and generates up to 1288 times fewer nodes than A* search. 

\end{abstract}

\section{Introduction}
\label{}
A* search \cite{hart1968formal} is an algorithm that searches for a sequence of actions that forms a path between a given start state and a given goal, where a goal is a set of goal states. By maintaining a search tree consisting of nodes that represent states and edges that represent transitions between states, search is performed by expanding nodes in this search tree, where nodes are prioritized for expansion according to a given cost. The expansion of a node and the computation of the cost of its child nodes is one of the most time-consuming portions of A* search. Node expansion is performed by applying every possible action to the state associated with a given node to \emph{generate} the child nodes. The cost of a node is computed by summing its path cost and heuristic value, where the path cost is the sum of transition costs from the start node to the given node and the heuristic value is computed by a heuristic function that estimates the cost to go from the state associated with the node to a closest goal state, commonly referred to as the cost-to-go.
Each iteration of A* search involves several steps: removing a node from the priority queue, expanding it, computing costs for its children, and pushing those children into the queue. As a result, the number of new nodes generated, heuristic function applications, and nodes pushed to the priority queue all grow linearly with the size of the action space. This escalating computational load can be substantial, particularly considering that the application of heuristic functions can be computationally expensive. Moreover, in numerous domains, the process of generating states can also be time-consuming, notably in motion planning and chemical synthesis. However, it is worth noting that much of this computational effort might be redundant, as A* search typically does not expand every single node it generates.

The need to reduce this linear increase in computational cost has become more relevant with the more frequent use of deep neural networks (DNNs) \cite{schmidhuber2015deep} as heuristic functions. While DNNs are universal function approximators~\cite{hornik1989multilayer}, they are computationally more expensive than heuristic functions that leverage domain knowledge, human intuition, or simplify the original problem to make it easier to solve~\cite{culberson1998pattern,MostowP89,bonet2001planning}.
Nonetheless, DNNs are able to learn heuristic functions to solve problems ranging from puzzles \cite{chen2011using, arfaee2011learning, deepcube, agostinelli2019solving}, to quantum computing \cite{zhang2020topological,bao2024twisty}, to chemical synthesis \cite{chen2020retro}, while making very few assumptions about the structure of the problem. Due to their flexibility and ability to generalize, DNNs offer the promise of learning heuristic functions in a largely domain-independent fashion. Removing the linear increase in computational cost as a function of the size of the action space would make DNNs practical for a wide range of applications with large action spaces, such as multiple sequence alignment, theorem proving, program synthesis, and chemical synthesis.

In this paper, we introduce \searchalg{}\footnote{The core idea behind Q* search was first posted on arXiv in February 2021.}, a search algorithm that makes use of state-action estimations (often known as Q-values) to reduce the number of node generations and heuristic function calls.
In each iteration of \searchalg{}, the number of nodes generated and number of heuristic function applications is independent of the size of the action space. \searchalg{} uses heuristic functions that map a single state to the  transition cost and the heuristic value for each of its successor states. This enables us to generate at most one node per iteration by storing tuples of nodes and actions in a priority queue, where priorities are determined by the estimated $f$-cost---i.e., the sum of the path cost from the start state, the transition cost estimation, and the cost-to-go estimation---analogous to A*, but without explicitly generating the successor nodes.
When removing a tuple of a node and action from the queue, we can then generate a new node by applying the action to the state associated with that node. As a result, the only aspect of \searchalg{} search that depends on the action space is pushing a node, along with each of the possible actions that can be applied to it, to the priority queue. This is also more memory efficient than explicitly generating all child nodes as, in our implementation, each action is only an integer. Our theoretical results show that \searchalg{} is guaranteed to find a shortest path given a heuristic function that does not overestimate the sum of the transition cost and cost-to-go of the state. 

Several techniques have been proposed to reduce the overhead of node generation and heuristic evaluation in heuristic search~\cite{yoshizumi2000partial, felner2012partial, helmert2006fast, richter2009preferred, GoldenbergFSHS13}. However, to the best of our knowledge, our method is the first to (1) avoid generating any successor nodes unless they are selected for expansion, and (2) require only a single heuristic function evaluation per expansion step. These properties enable our approach, Q*, to dramatically reduce computational and memory costs, particularly in domains with high branching factors or expensive heuristics. A detailed comparison with prior work is provided in Section~\ref{ssec:related}.

To obtain such state-action estimates, we learn both transition costs and cost-to-go values directly through domain interaction, without relying on any prior data. To this end, we extend the widely used method of deep approximate value iteration, which has previously been applied to learn state-based heuristic estimations and successfully combined with A* to solve complex problems~\cite{agostinelli2019solving}. Specifically, inspired by Q-learning~\cite{watkins1992q}, we employ a Deep Q-Network (DQN)~\cite{mnih2015human}, together with a tailored training scheme designed to produce accurate state-action cost estimates for use in our search algorithm. While learned heuristics may lead to overestimation and, therefore, do not guarantee optimality, they often produce optimal or near-optimal solutions in practice, and scale significantly better than traditional methods for computing heuristics~\cite{agostinelli2019solving}.

To evaluate \searchalg{} with our DQN-based heuristic, we conducted experiments across several challenging domains, including the Rubik’s Cube, Lights Out, and the 35-Pancake puzzle. The results demonstrate that \searchalg{} is orders of magnitude faster than A*, while also generating significantly fewer nodes and requiring far fewer heuristic evaluations.



\section{Background and Related Work}
A search problem instance, denoted as $I=(G, c, \start, \goal,h)$, is comprised of a graph $G=(V,E)$ with states (vertices) in $V$ and edges (transitions) in $E\subseteq V\times V$, and a transition cost function $c: E \rightarrow \mathbb{R}^{+}$ that assigns costs to graph edges. The instance specifies a starting state (\start) and a target state (\goal) or a predicate $P: V \rightarrow {0,1}$ indicating whether a state satisfies goal conditions. $h$ is a heuristic function which assigns to each state $s$ an estimate of the cost associated with the shortest path leading from $s$ to a nearest goal state, often termed the ``cost-to-go''. $h$ is called an\emph{admissible} heuristic function if never overestimates the cost-to-go for every state. 
The primary objective in heuristic search is to discover a path within graph $G$ that connects \start to \goal. The cost of the derived path is the cumulative cost of its constituent edges, determined by the transition cost function. We denote by $d(s,s')$ the shortest (cheapest) path between $s$ and $s'$ in $G$, and $d(\start,\goal)$ by $C^*$.

Note that the graph is typically given implicitly, where only the initial state is given alongside a set of transition functions $\mathcal{A}$. These functions represent various transitions, such as those between different robot configurations, puzzle permutations, or STRIPS-like states in domain-independent planning problems. In this work, we adhere to this assumption, meaning that we are provided with an \emph{action space} $\mathcal{A}$ (where $|\mathcal{A}|$ is often referred to as the branching factor), and we assume that the set of edges $E$ corresponds to applying each action $a \in \mathcal{A}$ from every state $s$. We denote the state resulted by applying action $a$ from state $s$ by $A(s,a)$ and the corresponding transition cost by $c^a(s)$. We use the term \emph{node} to refer to a data structure corresponding to a state, typically augmented with additional information such as cost-to-come (\(g\)), heuristic estimate (\(h\)), and possibly a pointer to its parent.

Typical search algorithms are primarily concerned with the number of node expansions, each of which involves generating all successors of a node and computing heuristic estimates for the resulting states. In this work, we decompose this process and focus specifically on the number of \emph{state generations}---i.e., the application of actions to states to produce successors---and the associated \emph{heuristic evaluations}, i.e., the number of calls to the heuristic function. Our goal is to reduce the number of state generations and to make heuristic estimation more efficient.

\subsection{A* and Batch Weighted A* Search}
A* search \cite{hart1968formal} is one of the most widely recognized and influential search algorithms. A* search maintains a priority queue, OPEN, from which it iteratively removes and expands the node with the lowest cost and a dictionary, CLOSED, that maps states that have already been generated to their path costs. 
The cost of each node is $f(n) = g(n) + h(n.s)$, where $g(n)$ is the path cost, the sum of transition costs along the path from \start to $n$, and $h(n.s)$ is the heuristic value, the estimated cost-to-go from the state associated with $n$ to a nearest goal state. After a node is expanded, its children whose states are not already in CLOSED have their states added to CLOSED and then pushed to OPEN. If the state of a child node $n$ is already in CLOSED, but the path cost of $n$ is cheaper than the path cost recorded in CLOSED, then the path cost of the state associated with $n$ is updated in CLOSED and $n$ is added to OPEN. The algorithm starts with only $n_{\start}$, a node associated with the \start state, in OPEN and terminates when a node associated with a goal state is removed from OPEN. 
A* search is guaranteed to find a shortest path if the heuristic function is admissible \cite{hart1968formal}.

When executing A* with a learned heuristic function (e.g., using deep neural networks), computing heuristic values can make A* search computationally expensive. To alleviate this issue, one can take advantage of the parallelism provided by graphics processing units (GPUs) by expanding the $B$ lowest cost nodes and computing their successors' heuristic values in parallel. Furthermore, even with a computationally cheap and informative heuristic, A* search can be both time and memory-intensive. To address this, one can trade potentially more costly solutions for potentially faster runtimes and less memory usage with a variant of A* search called weighted A* search \cite{pohl1970heuristic}. Weighted A* search computes the cost of each node as $f(n) = \lambda g(n) + h(n.s)$ where $\lambda \in [0, 1]$ is a scalar weighting. Weighted A* search is guaranteed to find a bounded suboptimal path if the heuristic function is admissible. The combination of expanding $B$ nodes every iteration and weighting the path cost by $\lambda$ is referred to as batch-weighted A* search (BWAS).  BWAS is a generalization of A* search, since A* search can be recovered by setting $\lambda$ to 1 and $B$ to 1. Crucially, when employing batch expansion in BWAS, the algorithm does not necessarily terminate immediately upon expanding a goal node. Instead, it maintains and refines upper and lower bounds on the optimal solution cost. The search proceeds until these bounds converge or a specified termination condition is met. This mechanism is essential for BWAS to guarantee bounded suboptimality, assuming the use of an admissible heuristic function~\cite{agostinelli2021obtaining, li2022optimal}. Pseudocode for the BWAS algorithm is presented in Algorithm~\ref{alg:bwas}. Black text corresponds to the standard A* algorithm, red highlights the modifications required for Weighted A*, and blue indicates the additions for expanding a batch of nodes at each iteration.

\begin{algorithm}[H]
	\footnotesize
	\caption{Batch Weighted A* Search (BWAS)}
	\label{alg:bwas}
	\begin{algorithmic}[1]
		\STATE {\bfseries Input:} $\start$, DNN $\ctgfunc$, batch size \textcolor{blue}{$B$}, weight \textcolor{red}{$\lambda$}
		\STATE OPEN $\gets$ priority queue of nodes based on minimal $f$
		\STATE CLOSED $\gets$ maps states to their shortest discovered path costs \\
		\STATE $UB, n_{UB} \leftarrow \infty, \text{NIL}$\\
		\STATE $LB \leftarrow 0$
		\STATE $n_{\start}\leftarrow \textrm{NODE}(s=\start, g=0,p=\text{NIL},f=\ctgfunc(\start))$
		\STATE PUSH $n_{\start}$ to OPEN \\
		\WHILE{not $\textrm{IS\_EMPTY}(\textrm{OPEN})$}
		\STATE generated $\leftarrow$ []
		\STATE batch\_expansions $\leftarrow 0$
		\WHILE{not $\textrm{IS\_EMPTY} (\textrm{OPEN})$ \textcolor{blue}{and $\textrm{batch\_expansions}< B$}}
		\STATE $n=(s, g,p,f)\leftarrow \textrm{POP(OPEN)}$
		\STATE batch\_expansions $\leftarrow$ batch\_expansions $ + 1$
		\IF{\textrm{IS\_EMPTY} (\textrm{generated})}
		\STATE $LB \leftarrow \max(f,LB)$
		\ENDIF
		\IF{$\textrm{IS\_GOAL}(s)$}
		\IF{$UB > g$}
		\STATE $UB,n_{UB} \leftarrow g,n$
		\ENDIF
		\STATE \textbf{continue loop}
		\ENDIF
		\FOR{$a$ in $|\mathcal{A}|$}
		\STATE $s' \leftarrow A(s,a)$
		\STATE $g' \leftarrow g + c^{a}(s)$
		\IF{$s'$ not in CLOSED or $g' < \textrm{CLOSED}[s']$}
		\STATE $\textrm{CLOSED}[s'] \leftarrow g'$
		\STATE $\textrm{APPEND} (\textrm{generated},(s',g',n))$
		\ENDIF
		\ENDFOR
		\ENDWHILE
		\IF{$LB \geq \textcolor{red}{\lambda \cdot} UB $}
		\STATE \textbf{return} $\textrm{PATH\_TO\_GOAL}(n_{UB})$
		\ENDIF
		\STATE generated\_states $\leftarrow \textrm{GET\_STATES}(\textrm{generated})$ 
		\STATE $\textrm{heuristics} \leftarrow \ctgfunc(\textrm{generated\_states})$
		\FOR{$0 \leq i \leq \textrm{SIZE}(\textrm{generated})$}
		\STATE $s,g,p \leftarrow \textrm{generated}[i]$
		\STATE $h \leftarrow \textrm{heuristics}[i]$
		\STATE $n_{s}\leftarrow \textrm{NODE}(s, g,p,f=\textcolor{red}{\lambda \cdot}g+h)$
		\STATE PUSH $n_{s}$ to OPEN
		\ENDFOR
		\ENDWHILE
		\STATE \textbf{return} $\textrm{PATH\_TO\_GOAL}(n_{UB})$ \COMMENT{failure if $n_{UB}$ is NIL}
	\end{algorithmic}
\end{algorithm}

\subsection{Learning Heuristic Functions}
Heuristic functions are central to search and planning, as accurate heuristics can drastically reduce the number of nodes an algorithm must expand to find a shortest path. Such functions may be derived manually using domain knowledge or generated automatically through algorithmic techniques. In grid navigation domains, for example, the Manhattan distance provides a simple, domain-specific lower bound on true path cost. More generally, heuristics can be computed by transforming the original problem into a simplified version that is easier to solve. Prominent examples include pattern databases (PDBs)~\cite{culberson1998pattern}, syntactic transformations~\cite{MostowP89}, and delete relaxations~\cite{bonet2001planning}. These methods often provide theoretical guarantees, but they also involve trade-offs between heuristic accuracy and computational cost. As problem complexity increases, the resources required to compute effective heuristics often grow prohibitively, resulting in either coarse approximations or limited scalability~\cite{agostinelli2019solving, muppasani2023solving}.

To address these limitations, recent work has turned to learning-based methods for constructing heuristic functions, aiming to automatically acquire heuristics from data without using domain-specific knowledge. Learned heuristics generally lack admissibility guarantees, meaning they cannot ensure that the solutions found are optimal. While recent work has explored ways to obtain admissible heuristics using deep neural networks~\cite{ernandes2004likely, agostinelli2021obtaining, li2022optimal}, this remains an active area of research. Despite the absence of formal guarantees, learned heuristics often perform well in practice---frequently producing optimal or near-optimal solutions with significantly fewer node expansions. Applications of learned heuristics span a range of challenging domains, including the Rubik's Cube~\cite{deepcube, agostinelli2019solving}, chemical synthesis~\cite{chen2020retro}, quantum circuit compilation~\cite{zhang2020topological, bao2024twisty}, theorem proving~\cite{kaliszyk2018reinforcement}, and robotics~\cite{tianmodel, eysenbach2019search}, where they have enabled solvers to find optimal or near-optimal solutions without human-engineered heuristics.

The idea of learning heuristics is not new; early work explored techniques for approximating existing heuristics or improving them through experience. One approach is imitation learning, which trains models to regress to cost-to-go values provided by an oracle or a domain-specific solver. For example, \citet{samadi2008compressing} trained neural networks to mimic PDB heuristics while reducing memory requirements. Other studies introduced neural architectures tailored for planning~\cite{abs-2112-01918, TakahashiSTW19, ShenTT20, FerberH020, ToyerTTX20} and designed alternative loss functions to promote effective search guidance~\cite{GarrettKL16, BhardwajCS17, GroshevGTSA18, abs-2310-19463}. These techniques often rely on high-quality demonstrations, which may not be available in complex or novel domains.

A complementary direction leverages reinforcement learning (RL), using search outcomes as feedback to iteratively refine heuristic functions. In early work, \citet{Bramanti-GregorD93} applied A* to generate training data, then used linear regression to improve future estimates. \citet{Fink07} extended this approach by learning weighted combinations of admissible heuristics. Later, \citet{ArfaeeZH11} incorporated neural networks and introduced curriculum learning via random walks to handle hard instances. More recently, \citet{OrseauL21} proposed learning both heuristic estimates and search policies simultaneously. Despite their promise, RL-based methods face challenges with sample inefficiency: they typically learn only from successful solutions, ignoring the many nodes expanded en route or during failed attempts. This bottleneck limits their ability to generalize and scale across problem instances.

\subsubsection{Learning Heuristic via Deep Approximate Value Iteration}
The learning approaches discussed above rely either on precomputed datasets of optimal solutions or on solving problems during search, collecting training data only when a solution is found. However, in complex domains, it is often impractical to obtain optimal solutions in advance or to solve problem instances without already having access to a strong heuristic. Deep Approximate Value Iteration (DAVI) addresses these challenges by neither requiring a dataset of optimal solutions nor depending on solved instances. Instead, it learns heuristic functions directly through value iteration, enabling learning even in domains where optimal solutions are scarce or expensive to compute.
Value iteration \cite{puterman1978modified} is a dynamic programming algorithm that is central in solving Markov Decision Processes (MDPs) and reinforcement learning problems~\cite{bellman1957dynamic,bertsekas1996neuro,sutton1998reinforcement}. It iteratively computes the expected value for each state, initially assuming a zero value for all states and progressively refining these estimations by applying a Bellman update. Value iteration is typically formulated for maximization problems featuring stochastic action effects and a potentially infinite planning horizon. However, in the realm of heuristic search, it can be redefined to address minimization problems, specifically aimed at minimizing costs, with deterministic effects and a finite planning horizon.
This redefinition is expressed by the following equation:
\begin{equation}
	\label{eq:viUpdateSimp}
	V'(s) = \min_{a \in \mathcal{A}}(c^a(s) + V(A(s,a)))
\end{equation}
In this context, $V$ represents a current estimate of the cost-to-go function, which approximates the cost of reaching the closest goal state from a given state. Equation (1) describes how to compute $V'(s)$, a refined and potentially more accurate estimate for the cost from state $s$, by incorporating the immediate cost and the estimated cost from successor states. This process can be viewed as a single step of value iteration, effectively improving an existing heuristic function. The resulting function $V'(s)$, after one or more such refinement steps, can then seamlessly serve as an effective heuristic function for A* search.





	
	
	Nonetheless, representing $V$ as a lookup table is too memory-intensive for problems with large state spaces. For instance, the Rubik's cube has $4.3 \times 10^{19}$ possible states. Therefore, we turn to approximate value iteration \citep{bertsekas1996neuro} where $V$ is represented as a parameterized function, $\ctgfunc$, with parameters $\ctgparams$. We choose to represent $\ctgfunc$ as a deep neural network (DNN). The parameters $\ctgparams$ are learned by using stochastic gradient descent to minimize a loss function. For a specific state $s$, the loss is computed as:
	\begin{equation}
		\label{eq:approxVi}
		L(\ctgparams, s) = \left(\min_{a \in \mathcal{A}}(c^a(s) + v_{\ctgparams^-}(A(s,a))) - \ctgfunc(s)\right)^2
	\end{equation}
	Where $\ctgparams^-$ are the parameters for the ``target'' DNN that is used to compute the updated cost-to-go. The term ``target'' arises because $v_{\ctgparams^-}(A(s,a))$ (and consequently the entire $\min$ term) provides a temporarily stationary target value for the loss function. This mechanism, originating from deep Q-networks \cite{mnih2015human}, decouples the computation of the target from the current network being optimized, which has been shown to result in a more stable training process. The parameters $\ctgparams^-$ are periodically updated to $\ctgparams$ during training. While we cannot guarantee convergence to $v_*$, approximate value iteration has been shown to approximate $v_*$ \cite{bertsekas1996neuro}. During training, this loss is minimized over a batch of sampled states from the environment.
	This combination of deep neural networks and approximate value iteration is referred to as deep approximate value iteration (DAVI). 
	
	The DeepCubeA algorithm~\cite{deepcube, agostinelli2019solving} is a prominent application of DAVI that successfully solves instances across multiple domains. It begins by training a heuristic function through sampling states by applying between 0 and $K$ random moves from a known goal state. The heuristic is then trained to minimize the loss defined in Equation~\ref{eq:approxVi} over these sampled states for a fixed number of iterations. To solve new problem instances, DeepCubeA uses the learned heuristic with BWAS. DeepCubeA was the first machine learning-based method to solve several challenging combinatorial puzzles while finding optimal solutions in the vast majority of verifiable cases, including the Rubik's Cube, the 24-puzzle, and Lights Out. Its success has since inspired extensions to domains such as quantum computing~\cite{zhang2020topological, bao2024twisty}, cryptography~\cite{jin20203d}, and parking lot optimization~\cite{siddique2021puzzle}.

	\subsection{Partial Expansions of States} \label{ssec:related}
	In A* (and most of its variants), expanding a node~$n$, which corresponds to state~$n.s$, involves the following steps:
	\begin{enumerate}
		\item Generating all successors of $n.s$.
		\item For each successor state $s'$:
		\begin{enumerate}
			\item Computing the heuristic value $h(s')$.
			\item Creating a node $n'$ corresponding to $s'$ and inserting it into \texttt{OPEN}, provided it has not already been discovered via a smaller or equal path cost.
		\end{enumerate}
	\end{enumerate}
	These operations can be very expensive, particularly when the branching factor is large or the heuristic function is computationally expensive. As a result, prior work in the heuristic search literature has explored partial expansions as a means of reducing this overhead.
	
	Partial expansion A* search (PEA*) \cite{yoshizumi2000partial} was proposed for problems with large branching factors. PEA* first expands a node by generating all of its children, however, it only keeps the children whose cost is below a certain threshold. It then adds a bookkeeping structure to remember the highest cost of the discarded nodes. The intention of PEA* is to save memory, however, the computational requirements do not reduce as every node removed from the priority queue has to be expanded and the heuristic function has to be applied to all of its children. Notably, PEA* is orthogonal to Q* and can be applied in conjunction with it to further reduce memory consumption.
	Enhanced partial expansion A* search (EPEA*) \cite{felner2012partial} uses a domain-specific knowledge function, called Operator  Selection  Function (OSF), to generate only a subset of children based on their cost, thus reducing the overall \emph{node generations}. In our work, we do not assume the availability of such domain knowledge.
	
	The EPEA* paper also introduces the concept of $\Delta$-PDB, which is highly relevant to our work. A standard pattern database (PDB) stores the distance from an abstract state to the abstract goal. When a heuristic value is needed for a state~$s$, it is retrieved by looking up $\text{PDB}[\phi(s)]$, where $\phi(s)$ denotes the abstraction of state~$s$. 
	In contrast, $\Delta$-PDB is defined over a state~$s \in V$ and an action~$a \in \mathcal{A}$, and it captures the change in heuristic value resulting from applying action~$a$ in state~$s$. Specifically, it represents the difference in distance to the abstract goal between the abstract state of~$s$ and the abstract successor state obtained by applying~$a$ to~$s$. Formally:
	\[
	\Delta\text{-PDB}[\phi(s),a] = \text{PDB}[\phi(s)] - \text{PDB}[\phi(A(s,a))],
	\]
	where $A(s,a)$ denotes the successor state resulting from applying action~$a$ to state~$s$. Equipped with $\Delta$-PDB, when given a node~$n$ corresponding to state~$n.s$, EPEA* computes $\Delta\text{-PDB}[\phi(n.s), a]$ for all actions~$a \in \mathcal{A}$ and generates only those successor nodes that would result in the same $f$-value as~$n$. While our approach also relies on evaluating the $f$-values of successors without explicitly generating them, it further delays node generation until the point of expansion.
	
	Another related work~\cite{GoldenbergFSHS13} extends the optimality analysis of A*~\cite{dechter1985generalized} from node expansions to node generations. It also introduces simple extensions of EPEA, called OGA* and SOGA*, which leverage additional domain knowledge to reduce node generations. Under certain definitions of optimality, these algorithms are shown to generate the minimal number of nodes with $f < C^*$ required to find an optimal solution. However, they may still generate nodes with \( f = C^* \) that are never expanded, and may require the same number of heuristic calls as A*, or even more in some cases.
	
	Deferred heuristic evaluation, introduced in the Fast Downward planner~\cite{helmert2006fast}, is a technique designed to reduce the number of heuristic evaluations in A* search. Typically, when a node is expanded and its successors are generated, the heuristic value of each successor is immediately computed. In contrast, deferred heuristic evaluation postpones this computation: each successor is inserted into the priority queue using its parent’s heuristic value, and its own heuristic is only computed when it is removed from the queue. This can significantly reduce the number of heuristic evaluations, but at the cost of reduced accuracy---particularly when the cost-to-go from a child node differs substantially from that of its parent. Moreover, this technique does not reduce the number of generated successors, but only the number of heuristic evaluations.
	We compare our proposed method, Q*, with deferred heuristic evaluation in our experiments and show that, in the vast majority of cases, Q* finds lower-cost solutions and does so significantly faster. Additionally, we observe that deferred heuristic evaluation occasionally runs out of memory due to its inability to prioritize among a node’s successors effectively.
	
	The Fast Downward planner introduced another relevant method: preferred operators (a term coined in the spirit of the ``helpful actions'' used in the Fast Forward planner~\cite{hoffmann2001ff}). These actions are used to prioritize promising successors during search, guiding the algorithm toward states that are more likely to lead to the goal, and are often given precedence in node selection to improve search efficiency and potentially generate fewer nodes. Preferred operators have also been combined with deferred heuristic evaluation to further reduce the number of heuristic computations~\cite{richter2009preferred}. While this approach is related to our work, there are several key differences. First, preferred operators partition actions into two categories—preferred and non-preferred—but do not provide a ranking among actions within each category. As a result, they lack the finer-grained prioritization needed for informed search. Second, preferred operators are typically derived from relaxed planning graphs, where actions selected in a solution to a relaxed version of the problem are marked as preferred. This method is well-suited to classical planning domains, such as those described in PDDL, but is not applicable to general graphs, which is the setting we consider in this work.

	
	
	Lazy A*~\cite{TolpinBSFK13,KarpasBSTF18} considers the setting where multiple heuristic functions are available, each with different computational costs and levels of accuracy. It reasons about when to evaluate each heuristic in order to reduce the overall search effort. While this approach is related to the general idea of partial expansions---since not all heuristics are evaluated at every expansion---it is orthogonal to our work, which focuses on the standard setting where a single heuristic function is used.
	
	\subsection{Q-learning}
	Instead of learning a function, $\ctgfunc$, that maps a state, $s$, to its cost-to-go, one can learn a function, $\dqnfunc$, that maps $s$ to its Q-factors, which is a vector containing $Q(s,a)$ for all actions $a$ \cite{bertsekas1995dynamic}. In a deterministic, finite-horizon environment, the Q-factor is defined as:
	
	\begin{equation}
		\label{eq:qfactor}
		Q(s,a) = c^a(s) + \gamma V(s')
	\end{equation}
	
	$V(s')$ can be expressed in terms of $Q$ with $V(s')=\min_{a'}{Q(s',a')}$. Learning $Q$ by iteratively updating the left-hand side of (\ref{eq:qfactor}) toward its right-hand side is known as Q-learning \cite{watkins1992q}. Like for DAVI, $Q$ is represented as a parameterized function, $\dqnfunc$, and we choose a deep neural network for $\dqnfunc$. This is also known as a deep Q-network (DQN) \cite{mnih2015human}. The architecture of the DQN is constructed such that the input is the state, $s$, and the output is a vector that represents $\dqnfunc(s,a)$ for all actions $a$. The parameters $\dqnparams$ are learned using stochastic gradient descent to minimize the loss function, which for specific state $s$ and action $a$, is defined as:
	
	\begin{equation}
		\label{eq:qlearn}
		L(\dqnparams,s,a) = \left((c^a(s) + \min_{a'\in \mathcal{A}}{q_{\dqnparams^-}(A(s,a),a')}) - \dqnfunc(s,a)\right)^2
	\end{equation}
	
	Just like in DAVI, the parameters $\dqnparams^-$ of the target DNN are periodically updated to $\dqnparams$ during training. 
	
	Similar to value iteration, Q-learning has been shown to converge to the optimal Q-factors, $q_*$, in the tabular case~\cite{watkins1992q}. In the approximate case, Q-learning has a computational advantage over DAVI because, while the number of parameters of the DQN grows with the size of the action space, the number of forward passes needed to compute the loss function stays constant for each update. We will show in our results that, in large action spaces, the training time for Q-learning is 127 times faster than DAVI.

	\section{\searchalg{} Search}
	We present \searchalg{}, a search algorithm that extends A* by leveraging state-action heuristic estimates rather than traditional state-based estimates. This approach reduces node generation and, to some extent, the number of heuristic evaluations. In A*, each node represents a state, whereas in \searchalg{}, nodes correspond to state-action pairs, which we refer to as \nodeacts{}. Specifically, a tuple \((s, a)\) represents the action \(a\) taken from state \(s\), and implicitly the resulting successor state. \searchalg{} utilizes two heuristic functions, in contrast to the single heuristic function \(h\) used by A*. The first, \(h_{c}(s,a)\), estimates the cost of applying action \(a\) in state \(s\), i.e. $c^{a}(s)$. The second, \(h_{d}(s,a)\), estimates the cost-to-go from \(A(s,a)\), the state resulting from that transition, i.e. $d(A(s,a),\goal)$.
	In various domains, transition costs ($c^{a}(s)$) for states are often predetermined and readily accessible, eliminating the need for estimation. However, in certain contexts, computing these costs can be resource-intensive. Take motion planning, for instance; calculating the transition cost between states (configurations) often requires executing a local planner, which can be computationally demanding. Moreover, transition costs may vary depending on the parent state in addition to the action, a concept known in planning as state-dependent action costs, adding another layer of complexity. For instance, in grid-like environments, the terrain of both the current and next states can influence the transition cost. Consequently, in the general case, \(h_{c}(s,a)\) approximates $c^{a}$, but this approximation can be substituted with the true transition cost if available.
	
	Given a state-action pair \((s,a)\), Let $s'$ be the successor state resulting from applying action \(a\) in state \(s\), i.e., \(A(s,a)\) . In A*, the successor state \(s'\) must be generated in order to compute its priority, where \(f(s') = g(s') + h(s')\). In contrast, \searchalg{} computes the priority of \((s,a)\) without generating \(s'\), using \(g(s) + h_c(s,a)\) as a substitute for \(g(s')\), and defining \(h_d(s,a) = h(A(s,a)) = h(s')\). The resulting priority function is:
	\[
	f(s,a) = g(s) + h_c(s,a) + h_d(s,a),
	\]
	which plays a similar role to A*'s \(f(s')\), while avoiding the explicit generation of \(s'\).  
	We refer to the estimated component of the priority function, \(h_c(s,a) + h_d(s,a)\), as the Q-factor, denoted \(Q(s,a)\). The transition cost and cost-to-go estimates are separated into two functions, rather than a single function estimating their sum. This separation enables support for bounded-suboptimal variants, where only the transition cost should be weighted---while the cost-to-go remains unscaled. Note that if the graph has uniform edge costs, or if only optimal solutions are sought, a single estimator for \(Q(s,a)\) suffices.
	
	We note that \(h_c(s,a)\) and \(h_d(s,a)\) are defined as two separate functions, where each function takes both a state and an action as input. Consequently, obtaining heuristic estimates for all successors of a given state requires invoking the two heuristic functions separately for each action, resulting in \(|\mathcal{A}|\) calls.
	However, it is often possible to define a heuristic function that, given a state, returns heuristic estimates for all of its successors in a single call. Formally, we define \(h_d(s) = (v_{a_1}, v_{a_2}, \ldots, v_{a_{|\mathcal{A}|}})\), where \(v_{a_i}\) denotes the cost-to-go estimate for the successor \(A(s, a_i)\). A similar formulation can be used for the transition cost function \(h_c\). Moreover, both functions can be merged into a single function $h_{cd}(s)$.
	This approach reduces the number of heuristic function calls from $|\mathcal{A}|$ to 1. While in some cases the computational cost of evaluating all successor heuristics at once is comparable to evaluating each successor individually, in many settings it is significantly more efficient to compute all values simultaneously. For instance, with pattern databases (PDBs), computing \(h_d(s)\) typically requires a single memory lookup, whereas computing each \(h_d(s,a)\) separately requires \(|\mathcal{A}|\) potentially non-sequential memory accesses. Similarly, in learning-based heuristics using neural networks, it is much more efficient to perform a single forward pass through the network on the state \(s\) to obtain all successor estimates, rather than invoking the network separately for each action.
	For simplicity, we continue to use the notations \(h_c(s,a)\) and \(h_d(s,a)\). From this point onward, these should be interpreted as retrieving the corresponding value from the output of \(h_{cd}(s)\), specifically the entry associated with action \(a\) and either the transition cost or the heuristic (corresponding to $h_c$ and $h_d)$.
	
	At every iteration, \searchalg{} search pops a \nodeact{}, $(s,a)$, from OPEN and generates a new state, $s'=A(s,a)$. Instead of explicitly generating all successors of \(s'\), \searchalg{} applies \(h_c(s')\) and \(h_d(s')\) to obtain the transition cost and cost-to-go estimates for all of its successor states in a single heuristic call. It is assumed that \(h_c\) and \(h_d\) only return values for applicable actions. Creating functions that operate in dynamic action spaces is discussed further in Section \ref{sec:discuss}. In \searchalg{} search, the only component that depends on the size of the action space is pushing nodes to OPEN. Unlike A*, which generates and evaluates all successors in each iteration, \searchalg{} generates only one node and applies the heuristic function just once per iteration.
	
	Similar to BWAS, we run \searchalg{} in batches of size \(B\) with a weight \(\lambda\), where \(B\) nodes are selected for expansion in each iteration. The goal is to find solutions with cost bounded by \(\lambda \times C^*\), yielding a weighted, batched variant we denote as \bwsearchalg{}.
	It is important to note that \searchalg{} is a special case of \bwsearchalg{}, where $\lambda = 1$ and $B = 1$. Consequently, the pseudocode for the \bwsearchalg{} algorithm, as outlined in Algorithm \ref{alg:bwqs}, inherently encompasses \searchalg{}. In the pseudocode, black text represents the standard \searchalg{} algorithm, red highlights the modifications for ensuring bounded suboptimality, and blue indicates the additions necessary for batch expansion at each iteration. Note that in the pseudocode, the start node is initialized with a special action \( a = \text{NO\_OP} \), which is not part of the domain's action set. This placeholder action indicates that applying \( a = \text{NO\_OP} \) to \start in line 15 of the algorithm results in the state \start itself.
	
	\begin{algorithm}[H]
		\footnotesize
		\caption{Batch Weighted \searchalg{} Search (\bwsearchalg{})}
		\label{alg:bwqs}
		\begin{algorithmic}[1]
			\STATE {\bfseries Input:} $\start$, state-action transition cost and cost-to-go heuristic $h_{cd}$, batch size \textcolor{blue}{$B$}, weight \textcolor{red}{$\lambda$}
			\STATE OPEN $\gets$ priority queue of nodes based on minimal $f$
			\STATE CLOSED $\gets$ maps states to their shortest discovered path costs \\
			\STATE $UB, n_{UB} \leftarrow \infty, \text{NIL}$\\
			\STATE $LB \leftarrow 0$
			\STATE $n_{\start}\leftarrow \textrm{NODE}(s=\start, g=0,p=\text{NIL},a=\text{NO\_OP}, f=0)$
			\STATE PUSH $n_{\start}$ to OPEN \\
			\WHILE{not $\textrm{IS\_EMPTY}(\textrm{OPEN})$}
			\STATE generated $\leftarrow$ []
			\STATE batch\_expansions $\leftarrow 0$
			\WHILE{not $\textrm{IS\_EMPTY} (\textrm{OPEN})$ \textcolor{blue}{and $\textrm{batch\_expansions}< B$}}
			\STATE $n=(s, a, g,p,f)\leftarrow \textrm{POP(OPEN)}$
			\STATE batch\_expansions $\leftarrow$ batch\_expansions + 1
			\IF{\textrm{IS\_EMPTY} (\textrm{generated})}
			\STATE $LB \leftarrow \max(f,LB)$
			\ENDIF
			\STATE $s' \leftarrow A(s,a)$
			\STATE $g' \leftarrow g + c^{a}(s)$
			\IF{$\textrm{IS\_GOAL}(s')$}
			\IF{$UB > g'$}
			\STATE $UB,n_{UB} \leftarrow g',n$
			\ENDIF
			\STATE \textbf{continue loop}
			\ENDIF
			\IF{$s'$ not in CLOSED or $g' < \textrm{CLOSED}[s']$}
			\STATE $\textrm{CLOSED}[s'] \leftarrow g'$
			\FOR{$a'$ in $|\mathcal{A}|$}
			\STATE $\textrm{APPEND} (\textrm{generated},(s',g',a', n))$
			\ENDFOR
			\ENDIF
			\ENDWHILE
			\IF{$LB \geq \textcolor{red}{\lambda \cdot} UB$}
			\STATE \textbf{return} $\textrm{PATH\_TO\_GOAL}(n_U)$
			\ENDIF
			\STATE generated\_states\_actions $\leftarrow \textrm{GET\_STATES}(\textrm{generated})$ 
			\STATE $\textrm{transition\_costs}, \textrm{heuristics} \leftarrow h_{cd}(\textrm{generated\_states\_actions})$
			\FOR{$0 \leq i \leq \textrm{SIZE}(\textrm{generated})$}
			\STATE $s,a, g,p \leftarrow \textrm{generated}[i]$
			\STATE $g' \leftarrow g + \textrm{transition\_costs}[i]$
			\STATE $h \leftarrow \textrm{heuristics}[i]$
			\STATE $n_{(s,a)}\leftarrow \textrm{NODE}(s,a, g,p,f=\textcolor{red}{\lambda \cdot} g'+h)$
			\STATE PUSH $n_{(s,a)}$ to OPEN
			\ENDFOR
			\ENDWHILE
			\STATE \textbf{return} $\textrm{PATH\_TO\_GOAL}(n_U)$ \COMMENT{failure if $n_U$ is NIL}
		\end{algorithmic}
	\end{algorithm}
	
	\section{Example Comparing \searchalg{} to Benchmark Algorithms}
	
	Figure~\ref{fig:example} illustrates the behavior of \searchalg{} during search, compared to A*, A* with deferred heuristics, and EPAE* with $\Delta$-PDB. The figure presents a problem instance where edge costs are shown next to the edges, along with the corresponding actions ($a_1$, $a_2$, or $a_3$). Each state displays its heuristic value. For \searchalg{}, which uses state-action estimations, we assume a perfect transition cost heuristic $h_c$ (i.e., edge costs are known without generating successor nodes) and that the cost-to-go estimate corresponds to the heuristic value of the resulting state, i.e., $h_d(s,a) = h(A(s,a))$. All algorithms are assumed to break ties in favor of smaller $h$-values, with secondary tie-breaking based on insertion order into the open list (a common default strategy).
	
	\searchalg{} begins by generating and expanding \start, invoking the heuristic function $h_{cd}$ on \start and adding $(\start,a_1)$, $(\start,a_2)$, and $(\start,a_3)$ to the open list with $f$-values 5, 3, and 3, respectively. It then pops $(\start,a_2)$, generating $v_2$, evaluating its heuristic, and inserting $(v_2,a_1)$, $(v_2,a_2)$, and $(v_2,a_3)$ into the open list with $f$-values 4, 3, and 3. Next, $(\start,a_3)$ is popped, generating $v_3$, calling the heuristic on it, and inserting $(v_3,a_1)$, $(v_3,a_2)$, and $(v_3,a_3)$ with $f$-values 4, 3, and 3. Finally, due to the tie-breaking rule, \goal is generated, and the algorithm terminates after 4 node generations and 4 heuristic calls.
	
	A* begins by generating \start and then generates $v_1$, $v_2$, and $v_3$, evaluating their heuristics and inserting them into the open list with $f$-values 5, 3, and 3. It then expands $v_2$, generating $v_7$, $v_8$, and $v_9$, evaluating their heuristics, and inserting them with $f$-values 4, 3, and 3. Afterwards, $v_3$ is expanded, generating $v_{10}$, $v_{11}$, and \goal, which are also evaluated and inserted with $f$-values 4, 3, and 3. Finally, \goal is expanded and the search terminates. In total, A* generates 10 nodes and performs 10 heuristic evaluations.
	
	A* with deferred heuristics reduces the number of heuristic evaluations but may expand more nodes due to deferred (and inaccurate) $h$-values. It begins similarly to A*, but upon expanding \start, the heuristics of $v_1$, $v_2$, and $v_3$ are not yet computed. Instead, they are inserted into the open list using the heuristic of \start (which is 3), resulting in $f=4$ for all. Then $v_1$ is expanded, generating $v_4$, $v_5$, and $v_6$ (which A* does not generate), inserted with $f=7$ using $v_1$'s heuristic of 4. Next, $v_2$ is expanded, generating $v_7$–$v_9$, inserted with $f$-values 5, 4, and 4, respectively. Then $v_3$ is expanded, generating $v_{10}$, $v_{11}$, and \goal, with $f$-values 5, 4, and 5. Since \goal has $f=5$ and was the last node inserted with that value, the algorithm must expand all nodes with $f \leq 5$ before reaching \goal. This requires evaluating the heuristics of $v_7$ through $v_{11}$. Overall, A* with deferred heuristics generates 13 nodes (3 more than A*) and performs 10 heuristic evaluations.
	
	Finally, EPAE* with $\Delta$-PDB generates and expands \start, calling the heuristic function on $(\start,a_1)$, $(\start,a_2)$, and $(\start,a_3)$. It generates $v_2$ and $v_3$ with $f=3$, and reinserts \start with $f=5$ to represent $v_1$ without generating it. Then, $v_2$ is expanded, evaluating $(v_2,a_1)$, $(v_2,a_2)$, and $(v_2,a_3)$, and generating $v_8$ and $v_9$ with $f=3$, as well as reinserting $v_2$ with $f=4$ (to account for $v_7$). Similarly, $v_3$ is expanded, evaluating $(v_3,a_1)$, $(v_3,a_2)$, and $(v_3,a_3)$, generating $v_{11}$ and \goal with $f=3$, and reinserting $v_3$ with $f=4$ (to account for $v_{10}$). Finally, \goal is expanded due to tie-breaking. In total, EPAE* generates 7 nodes and performs 10 heuristic evaluations.
	
	A summary table below the figure reports the total number of node generations and heuristic evaluations for each algorithm, highlighting the superior performance of \searchalg{}.

	\begin{figure}[tbph]
		\centering
		\begin{tikzpicture}[
			node/.style={circle, draw, minimum size=1.2cm, align=center},
			edge/.style={->, thick},
			label/.style={font=\footnotesize}
			]
			
			\node[node] (A) at (0,0) {\start\\h=3};
			
			\node[node] (B) at (-4.5,-2.5) {$v_1$\\h=4};
			\node[node] (C) at (0,-2.5)  {$v_2$\\h=2};
			\node[node] (D) at (4,-2.5)  {$v_3$\\h=2};
			
			\draw[edge] (A) -- 
			node[label, pos=0.2, anchor=east] {1} 
			node[label, pos=0.8, anchor=east] {$a_1$} 
			(B);
			\draw[edge] (A) -- 
			node[label, pos=0.2, anchor=east] {1} 
			node[label, pos=0.8, anchor=east] {$a_2$} 
			(C);
			\draw[edge] (A) -- 
			node[label, pos=0.2, anchor=west] {1} 
			node[label, pos=0.8, anchor=west] {$a_3$} 
			(D);
			
			\node[node] (G) at (-6,-5) {$v_4$\\h=2};
			\node[node] (H) at (-4.5,-5) {$v_5$\\h=2};
			\node[node] (I) at (-3,-5) {$v_6$\\h=2};
			
			\draw[edge] (B) -- 
			node[label, pos=0.2, anchor=east] {2} 
			node[label, pos=0.8, anchor=east] {$a_1$} 
			(G);
			\draw[edge] (B) -- 
			node[label, pos=0.2, anchor=east] {2} 
			node[label, pos=0.8, anchor=east] {$a_2$} 
			(H);
			\draw[edge] (B) -- 
			node[label, pos=0.2, anchor=west] {2} 
			node[label, pos=0.8, anchor=west] {$a_3$} 
			(I);
			
			\node[node] (J) at (-1.5,-5) {$v_7$\\h=1};
			\node[node] (K) at (0,-5)  {$v_8$\\h=1};
			\node[node] (L) at (1.5,-5)  {$v_9$\\h=1};
			
			\draw[edge] (C) -- 
			node[label, pos=0.2, anchor=east] {2} 
			node[label, pos=0.8, anchor=east] {$a_1$} 
			(J);
			\draw[edge] (C) -- 
			node[label, pos=0.2, anchor=east] {1} 
			node[label, pos=0.8, anchor=east] {$a_2$} 
			(K);
			\draw[edge] (C) -- 
			node[label, pos=0.2, anchor=west] {1} 
			node[label, pos=0.8, anchor=west] {$a_3$} 
			(L);
			
			\node[node] (M) at (3,-5)  {$v_{10}$\\h=1};
			\node[node] (N) at (4.5,-5)  {$v_{11}$\\h=1};
			\node[node] (O) at (6,-5)  {\goal\\h=0};
			
			\draw[edge] (D) -- 
			node[label, pos=0.2, anchor=east] {2} 
			node[label, pos=0.8, anchor=east] {$a_1$} 
			(M);
			\draw[edge] (D) -- 
			node[label, pos=0.2, anchor=east] {1} 
			node[label, pos=0.8, anchor=east] {$a_2$} 
			(N);
			\draw[edge] (D) -- 
			node[label, pos=0.2, anchor=west] {2} 
			node[label, pos=0.8, anchor=west] {$a_3$} 
			(O);
			
		\end{tikzpicture}
		
		\vspace{1em}
		
		\begin{tabular}{p{3cm} | p{4.8cm} | p{4.8cm}}
			\textbf{Algorithm} & \textbf{Nodes Generated} & \textbf{Heuristic Calls} \\
			\hline
			Q* & 4: \start, $v_2$, $v_3$, \goal & 4: \start, $v_2$, $v_3$, \goal \\
			\hline
			A* & 10: \start, $v_1$, $v_2$, $v_3$, $v_7$, $v_8$, $v_9$, $v_{10}$, $v_{11}$, \goal & 10: \start, $v_1$, $v_2$, $v_3$, $v_7$, $v_8$, $v_9$, $v_{10}$, $v_{11}$, \goal \\
			\hline
			Deferred \newline Heuristics & 13: \start, $v_1$, $v_2$, $v_3$, $v_4$, $v_5$, $v_6$, $v_7$, $v_8$, $v_9$, $v_{10}$, $v_{11}$, \goal & 10: \start, $v_1$, $v_2$, $v_3$, $v_7$, $v_8$, $v_9$, $v_{10}$, $v_{11}$, \goal \\
			\hline
			EPEA* \newline (with $\Delta$-PDB) & 7: \start, $v_2$, $v_3$, $v_8$, $v_9$, $v_{11}$, \goal & 10: \start, $v_1$, $v_2$, $v_3$, $v_7$, $v_8$, $v_9$, $v_{10}$, $v_{11}$, \goal \\
			\hline
		\end{tabular}
		
		\caption{Example demonstrating the node generations and heuristic calls by each algorithm.}
		\label{fig:example}
	\end{figure}

	\section{Theoretical Analysis}
	
	We will show that the \bwsearchalg{} algorithm qualifies as a bounded-suboptimal search approach. This indicates that it is ensured to discover a path with a cost $U \leq \frac{C^*}{\lambda}$, if such a path exists. This holds true under the condition that all Q-factors $Q(s,a)$ never overestimate $c^a(s)+d(A(s,a),\goal)$;\footnote{Note that each component can be overestimated individually, as long as the sum of both components does not result in an overall overestimation.} we refer to a heuristic function meeting this criterion as q-admissible. This proof is an adaptation of the proof that A* search is an admissible search algorithm \cite{hart1968formal}.
	\begin{lemmaT} \label{lemma:open}
		As long as \bwsearchalg{} did not terminate, either there exists a node in OPEN corresponding to a prefix of some shortest path from \start to \goal, or a shortest path from \start to \goal was discovered.
		
		\begin{proof}
			At the beginning of the search, $n_{(\start, \text{NO\_OP})}$ is in OPEN with\\ $g(n_{(\start, \text{NO\_OP})}) = 0$, which is the prefix of any shortest path from \start to \goal. In every search iteration $i>1$ if a shortest path from \start to \goal was not discovered, let $P$ be some shortest path from \start to \goal and $\Delta$ be the set of closed nodes in $P$, that were expanded with optimal $g$-value. That is, $\Delta = \{n | n\in P \text{ and }, n \in \text{ CLOSED } \text{ and } g(n)=d(\start,n)\}$. $\Delta$ is not empty, as after the first search iteration, $\start \in \Delta$. Let $n^*$ be the element in $\Delta$ with the highest index. Since an optimal path from \start to \goal has not been discovered, $n^* \neq \goal$. Let $n'$ be the successor of $n'$ in $P$. Due to the optimality of $P$, $g(n') = d(\start,n')$. In addition, since $n' \notin \Delta$ and $n^* \in \Delta$, $n'$ is in OPEN.
		\end{proof}
	\end{lemmaT}
	Using Lemma~\ref{lemma:open}, we prove our main theorem.
	
	\begin{theorem}
		Given that all transition costs are greater than zero, $0 \leq \lambda \leq 1$, and a q-admissible heuristic function, \bwsearchalg{} is bounded suboptimal. That is, \bwsearchalg{} returns a solution with a cost bounded by $\frac{1}{\lambda} \cdot C^*$, if such a solution exists.
	\end{theorem}
	
	\begin{proof}
		First, it is important to recognize that \bwsearchalg{} consistently maintains information about the shortest path discovered so far to the \goal, identified with a cost denoted as $U$. Upon termination, \bwsearchalg{} returns this solution. Termination of \bwsearchalg{} occurs under two conditions: either a solution is found with $\lambda \cdot U \leq LB$, or the OPEN set becomes empty after exhaustively expanding all nodes in the graph. Consequently, if there exist paths from \start to \goal, \bwsearchalg{} is guaranteed to discover one.
		
		Now, we aim to show that the path returned by \bwsearchalg{} is bounded by $\frac{1}{\lambda} \cdot C^*$. Assume by contradiction that \bwsearchalg{} has terminated and produced a solution with a cost $U > \frac{1}{\lambda} \cdot C^*$. As the algorithm has concluded, we have $LB \geq \lambda \cdot U$, indicating that at least one node was expanded with a priority greater than or equal to $\lambda \cdot U$. Let $n_{(s,a)}$ denote the first node expanded during the search with a priority greater than or equal to $\lambda \cdot U$. According to Lemma~\ref{lemma:open}, at the moment $n_{(s,a)}$ was chosen for expansion, there existed another node $n_{(s',a')}$ in OPEN, corresponding to an optimal path (costing $C^*$).
		
		Since $n_{(s,a)}$ was expanded instead of $n_{(s',a')}$, we infer that at the moment of expansion,
		\begin{equation} \label{eq:contradition}
			\lambda \cdot g(n_{(s,a)}) + \lambda \cdot h_{c}(s,a) + h_{d}(s,a) \leq \lambda \cdot g(n_{(s',a')}) + \lambda \cdot h_{c}(s',a') + h_{d}(s',a')
		\end{equation}
		By virtue of q-admissibility and $n_{(s',a')}$ being a node on an optimal path, it holds that 
		\[g(n_{(s',a')}) + h_{c}(s',a') +  h_{d}(s',a') \leq C^*\] 
		Since $\lambda \in [0,1]$, it also holds that 
		\[
		\lambda \cdot g(n_{(s',a')}) + \lambda \cdot h_{c}(s',a') +  h_{d}(s',a') \leq C^*
		\]
		Thus, due to Equation~\ref{eq:contradition},
		\begin{equation} \label{eq:contradition2}
			\lambda \cdot g(n_{(s,a)}) + \lambda \cdot h_{c}(s,a) + h_{d}(s,a) \leq C^*
		\end{equation}
		However, $n_{(s,a)}$ denotes the first node expanded during the search with a priority greater than or equal to $\lambda \cdot U$, therefore,
		\[\lambda \cdot g(n_{(s,a)}) + \lambda \cdot h_{c}(s,a) + h_{d}(s,a) \geq \lambda \cdot U\] 
		Since $U > \frac{1}{\lambda} \cdot C^*$, it holds that 
		\[\lambda \cdot g(n_{(s,a)}) + \lambda \cdot h_{c}(s',a') + h_{d}(s',a') > C^*\]
		In contradiction with Equation~\ref{eq:contradition2}. Consequently, if a solution exists, BWQS terminates and return a solution of cost $U \leq  \frac{1}{\lambda} \cdot C^*$.

	\end{proof}

	\section{Action-cost Heuristic Estimations via Deep Q-Learning}
	
	The transition cost and cost-to-go estimates, \(h_c\) and \(h_d\), can be obtained in various ways. One approach is to use a \(Q\)-PDB, which, given a state \(s\), returns a tuple of the form
	\[
	\text{Q-PDB}(\phi(s)) = \big(d(\phi(A(s,a_1)), \text{goal}), \dots, d(\phi(A(s,a_{|\mathcal{A}|})), \text{goal})\big),
	\]
	where each entry corresponds to the shortest path distance from the abstract successor state \(\phi(A(s,a))\) to the goal. This approach is inspired by \(\Delta\)-PDBs, but differs in that it requires only a single heuristic lookup (i.e., a single memory access) and stores the full values rather than the differences between parent and successor estimates---though a similar differential treatment could also be applied.
	In this work, however, we explore obtaining the transition cost and cost-to-go estimates using deep Q-learning.
	
	We train the state-based cost-to-go function, \( q_\phi \), parameterized by a neural network with weights~\( \phi \), using a procedure similar to that of DVAI~\citep{agostinelli2019solving}. A separate network is trained for each domain.
	To generate training data, we perform \( K \)-step random walks from the goal state. Each encountered state is added to a training batch of size \( TB \) (not to be confused with the search-time batch size \( B \) used by BWAS and BWQS after training). For each state \( s \in TB \), we sample one action according to a Boltzmann distribution:
	\begin{equation}
		\label{eq:boltzmann}
		p_{s,a} = \frac{e^{-\dqnfunc(s,a)/T}}{\sum_{a'=1}^{|\mathcal{A}|}e^{-\dqnfunc(s,a')/T}},
	\end{equation}
	where \( T = \frac{1}{3} \) is the temperature. This results in a batch of \( TB \) state-action pairs used for training.
	At each training iteration, we compute the loss (see Equation~\ref{eq:qlearn}) \emph{over the current training batch} and update \( \phi \) by minimizing this loss using the ADAM optimizer~\citep{kingma2014adam}. This data collection and optimization process is repeated for \( I \) training iterations. To stabilize learning, we employ a target network with parameters~\( \phi^- \), which is used to compute target values in the loss function. The target network is periodically updated to match the current network parameters~\( \phi \), following the update schedule used in the DeepCubeA implementation~\citep{deepcubeagithub}.
	
	We train the state-based cost-to-go function with the same network architecture described in \citet{agostinelli2019solving}, which has a fully connected layer of size 5,000, followed by another fully connected layer of size 1,000, followed by four fully connected residual blocks of size 1,000 with two hidden layers per residual block \cite{he2016deep}, followed by a layer of size 1 representing the cost-to-go. The state-action-based cost-to-go function (DQN, $\dqnfunc{}(s,a)$) also has the same architecture with the exception that the output layer is a vector that estimates the cost-to-go for taking every possible action. These architectures are visualized in Figure~\ref{fig:arch}.
	This implementation of $\dqnfunc{}(s,a)$ estimates $c^{a}(s) + h(A(s,a))$ as a single entity, which is sufficient for our evaluation domains, where all edge costs are uniform. 
	However, it is straightforward to modify the network architecture to separate the computation of transition costs and cost-to-go values using a shared backbone with two output heads: one that outputs a vector of transition costs (one per action), and another that outputs a vector of cost-to-go estimates (also one per action). This design allows for domains with non-uniform edge costs while maintaining efficiency in computing all Q-values in a single forward pass.

	\begin{figure}[th]
		\centering
		
		\begin{subfigure}[t]{\textwidth}
			\centering
			\begin{adjustbox}{max width=\textwidth}
				\begin{tikzpicture}[
					node distance=1.0cm and 1.0cm, 
					layer/.style={rectangle, draw, minimum width=3.6cm, minimum height=1.5cm, font=\large},
					resblock/.style={rectangle, draw, fill=gray!10, minimum width=4.2cm, minimum height=1.8cm, font=\large\linespread{0.9}\selectfont, align=center},
					arrow/.style={->, line width=1.2pt, >=latex}
					]
					
					\node[layer] (input) {Input};
					\node[layer, right=of input] (fc1) {FC (5000)};
					\node[layer, right=of fc1] (fc2) {FC (1000)};
					\node[resblock, right=of fc2] (rb1) {Residual Block 1\\(2 FC layers, size 1000)};
					\node[resblock, right=of rb1] (rb2) {Residual Block 2\\(2 FC layers, size 1000)};
					\node[resblock, right=of rb2] (rb3) {Residual Block 3\\(2 FC layers, size 1000)};
					\node[resblock, right=of rb3] (rb4) {Residual Block 4\\(2 FC layers, size 1000)};
					\node[layer, right=of rb4, align=center] (fc3) {FC (1)\\Cost-to-go};
					
					\draw[arrow] (input) -- (fc1);
					\draw[arrow] (fc1) -- (fc2);
					\draw[arrow] (fc2) -- (rb1);
					\draw[arrow] (rb1) -- (rb2);
					\draw[arrow] (rb2) -- (rb3);
					\draw[arrow] (rb3) -- (rb4);
					\draw[arrow] (rb4) -- (fc3);
				\end{tikzpicture}
			\end{adjustbox}
			\caption{DAVI: Predicts scalar cost-to-go.}
		\end{subfigure}
		
		\vspace{1.5em}
		
		\begin{subfigure}[t]{\textwidth}
			\centering
			\begin{adjustbox}{max width=\textwidth}
				\begin{tikzpicture}[
					node distance=1.0cm and 1.0cm, 
					layer/.style={rectangle, draw, minimum width=3.6cm, minimum height=1.5cm, font=\large},
					resblock/.style={rectangle, draw, fill=gray!10, minimum width=4.2cm, minimum height=1.8cm, font=\large\linespread{0.9}\selectfont, align=center},
					arrow/.style={->, line width=1.2pt, >=latex}
					]
					
					\node[layer] (input) {Input};
					\node[layer, right=of input] (fc1) {FC (5000)};
					\node[layer, right=of fc1] (fc2) {FC (1000)};
					\node[resblock, right=of fc2] (rb1) {Residual Block 1\\(2 FC layers, size 1000)};
					\node[resblock, right=of rb1] (rb2) {Residual Block 2\\(2 FC layers, size 1000)};
					\node[resblock, right=of rb2] (rb3) {Residual Block 3\\(2 FC layers, size 1000)};
					\node[resblock, right=of rb3] (rb4) {Residual Block 4\\(2 FC layers, size 1000)};
					\node[layer, right=of rb4, align=center] (fc3) {FC $(|\mathcal{A}|)$\\Cost-to-go per action};
					\draw[arrow] (input) -- (fc1);
					\draw[arrow] (fc1) -- (fc2);
					\draw[arrow] (fc2) -- (rb1);
					\draw[arrow] (rb1) -- (rb2);
					\draw[arrow] (rb2) -- (rb3);
					\draw[arrow] (rb3) -- (rb4);
					\draw[arrow] (rb4) -- (fc3);
				\end{tikzpicture}
			\end{adjustbox}
			\caption{Q-learning: Predicts cost-to-go for each action in \( \mathcal{A} \).}
		\end{subfigure}
		
		\caption{Comparison of DAVI and Q-learning architectures. Both share the same backbone but differ in their output layers.}
		\label{fig:arch}
	\end{figure}

	\section{Experimental Evaluation}
	\label{sec:results}
	
	\begin{figure*}[t]
		\centering
		\begin{subfigure}{0.3\textwidth}
			\centering
			\includegraphics[width=1\textwidth]{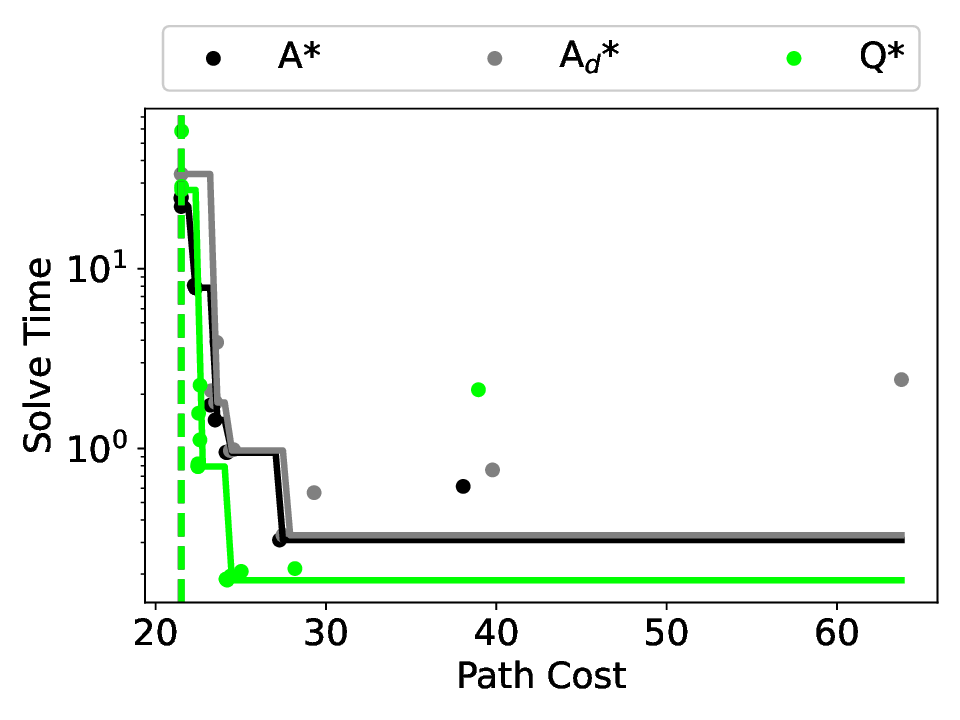}
			\caption{RC}
		\end{subfigure}
		\begin{subfigure}{0.3\textwidth}
			\centering
			\includegraphics[width=1\textwidth]{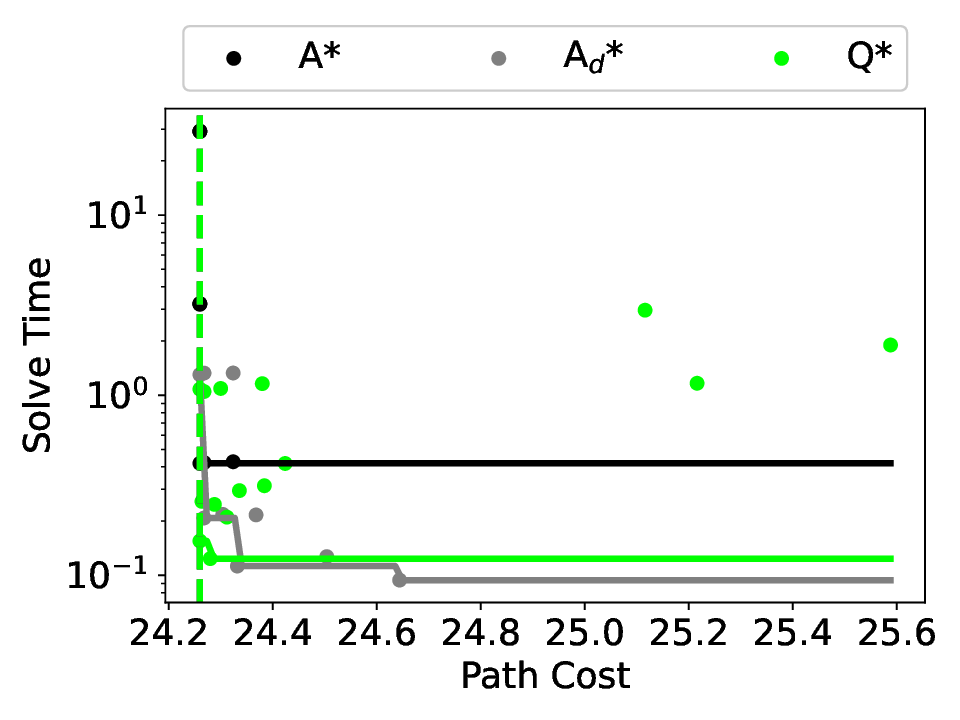}
			\caption{Lights Out}
		\end{subfigure}
		\begin{subfigure}{0.3\textwidth}
			\centering
			\includegraphics[width=1\textwidth]{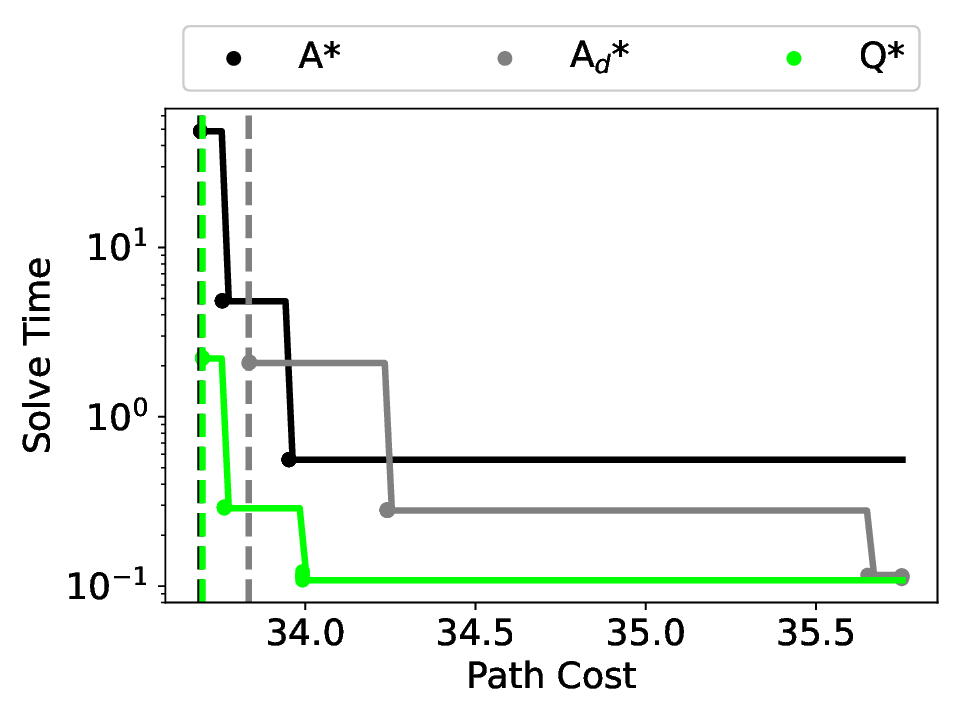}
			\caption{35-pancake}
		\end{subfigure}
		\caption{Relationship between the average path cost and the average time to find a solution.}
	\label{fig:searchtime}
\end{figure*}

\begin{figure*}[t]
	\centering
	\begin{subfigure}{0.3\textwidth}
		\centering
		\includegraphics[width=1\textwidth]{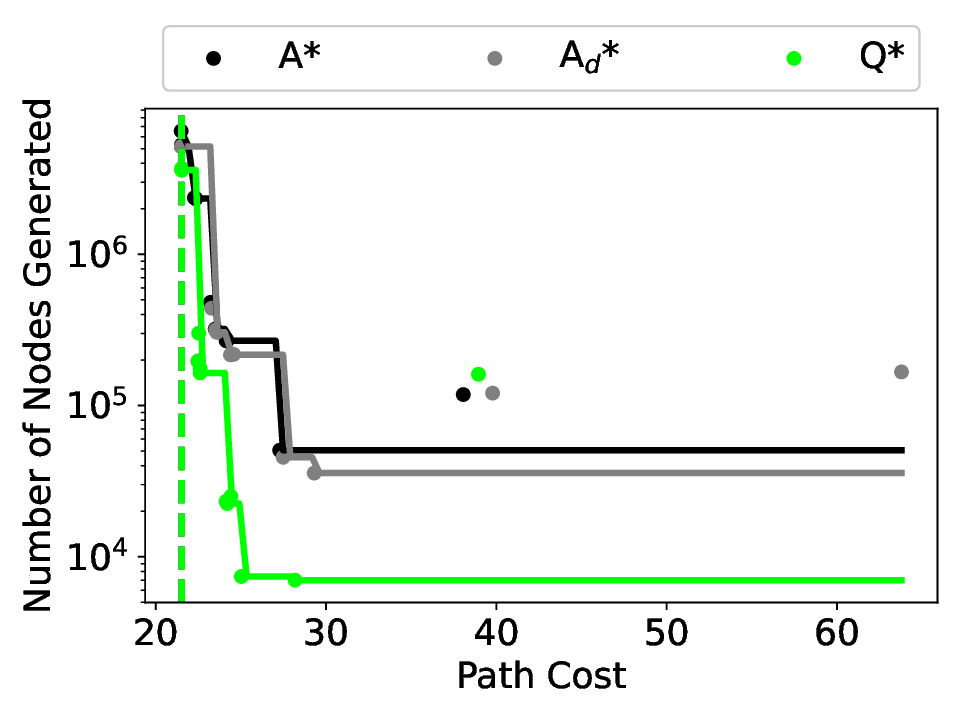}
		\caption{RC(12)}
	\end{subfigure}
	\begin{subfigure}{0.3\textwidth}
		\centering
		\includegraphics[width=1\textwidth]{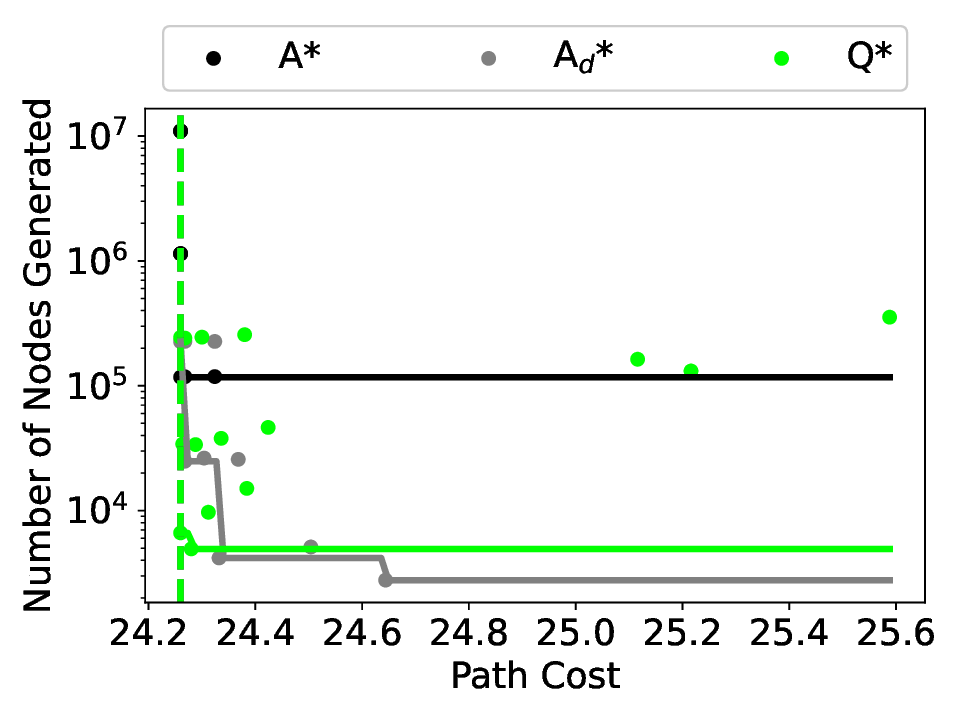}
		\caption{Lights Out}
	\end{subfigure}
	\begin{subfigure}{0.3\textwidth}
		\centering
		\includegraphics[width=1\textwidth]{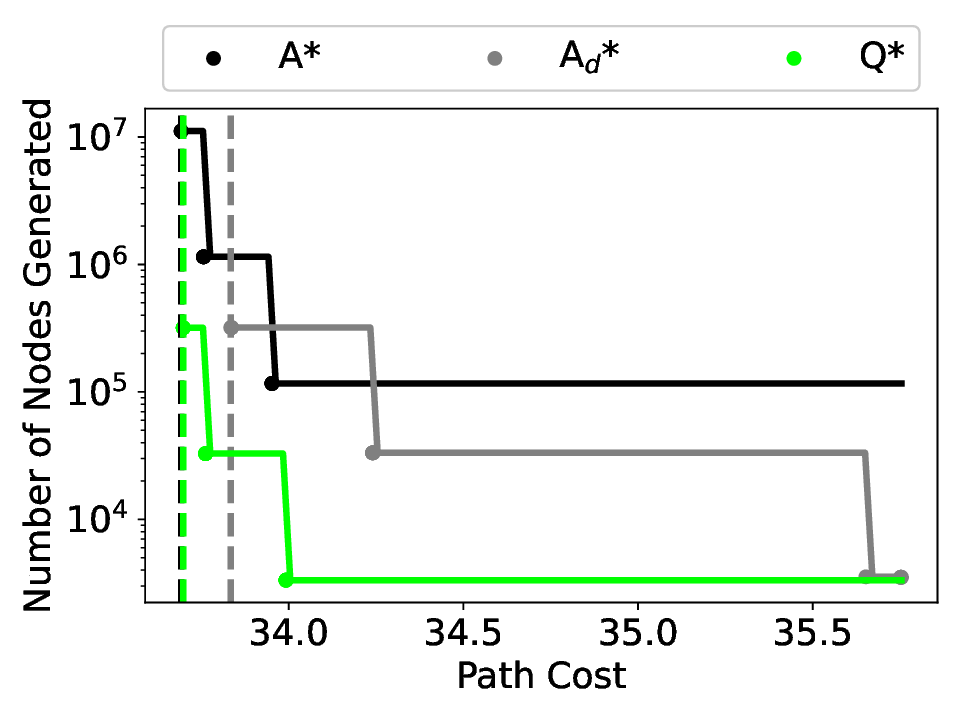}
		\caption{35-Pancake}
	\end{subfigure}
	\caption{Relationship between the average path cost and the average node generations.}
	\label{fig:searchnodes}
\end{figure*}
In this section, we detail our empirical evaluation of \searchalg{}.
\subsection{Settings, Baselines, and Network Architectures}
We evaluate \searchalg{} across various domains, including the Rubik's Cube (which has 12 actions), the 7 by 7 Lights Out puzzle \cite{agostinelli2019solving} (which has 49 actions), and the 35-Pancake puzzle (which has 49 actions). 

We compare \searchalg{} to both standard A* search and the deferred version of A* search~\cite{helmert2006fast}, which we refer to as A$_d$*. In A$_d$*, the heuristic value of each child is initially set to that of its parent, deferring the actual heuristic computation until the child is selected for expansion. In our implementation, we further optimize this approach by delaying the generation of successors until a node reaches the top of the priority queue, thereby reducing the number of state generations even further compared to the original deferred heuristic evaluation method. All algorithms expand a batch of nodes $N$, instead of a single node at each iteration, and use a weight of $\lambda$ (i.e., we evaluated the batch-weighted version for each algorithm). Both A* and A$_d$* employ a state-based cost-to-go function model trained using DVAI. In contrast, \searchalg{} utilizes a state-action-based cost-to-go estimation, trained using Q-learning.

For generating training states to train the heuristics, the number of times we scramble the puzzle, $K$, is set to 30 for the Rubik's cube, 50 for Lights Out, and 70 for the 35-Pancake puzzle. 
In addition, unless specified otherwise, we used a training batch size of $TB=10,000$ and $I=1.2$ million iterations, for both DAVI and Q-learning. The machines we use for training and search have 48 2.4 GHz Intel Xeon central processing units (CPUs), 192 GB of random access memory, and two 32GB NVIDIA V100 GPUs.

Prior work on solving the Rubik's cube with deep reinforcement learning and A* search used $\lambda=0.6$ and $N=10000$ for BWAS \cite{agostinelli2019solving}. However, since these search parameters create a tradeoff between speed, memory usage, and path cost, we also examine the performance with different parameter settings to understand how A* search and \searchalg{} search compare along these dimensions. Therefore, we try all combinations of $\lambda \in \{0.0,0.2,0.4,0.6,0.8,1.0\}$ and $N \in \{100, 1000, 10000\}$. For each method and each action space, we prune all combinations that cause our machine to run out of memory or that require over 24 hours to complete. We use the same 1,000 test states used for the Rubik's cube and 500 test states for Lights Out as used in previous work by \citet{agostinelli2019solving}. We generated 500 test states for the 35-Pancake puzzle. Each test state was obtained by scrambling the puzzle between 1,000 and 10,000 steps. We scramble the puzzle this many times because it is significantly larger than the number of steps it takes to solve these puzzles, which will hopefully result in states further away from the goal. For example, previous work scrambled the Rubik's cube and 15-puzzle this many times, which resulted in an average optimal cost-to-go of 20.6 and 52.0, respectively, while the cost of a longest shortest path was 26 and 80, respectively \cite{agostinelli2019solving}.

\subsection{Results}
\begin{table}[tb]
	\centering
	\caption{The table shows the number of training iterations per second with a training batch size of 10,000 and the projected number of days to train for 1.2 million iterations. DAVI is significantly slower than Q-learning, especially when the size of the action space is large.}
	\label{tab:traintime}
	\vskip 0.15in
	\begin{tabular}{@{}cccc@{}}
		\toprule
		Puzzle                    & Method & Itrs/Sec      & Train Time    \\ \midrule
		\multirow{2}{*}{RC(12)}   & DAVI    & 3.96          & 3.5d          \\
		& Q-learning    & \textbf{8.55} & \textbf{1.6d} \\ \midrule
		\multirow{2}{*}{RC(156)}  & DAVI    & 0.42          & 33d           \\
		& Q-learning    & \textbf{7.46} & \textbf{1.9d} \\ \midrule
		\multirow{2}{*}{RC(1884)} & DAVI    & 0.04          & 347d          \\
		& Q-learning    & \textbf{5.08} & \textbf{2.7d} \\ \bottomrule
	\end{tabular}
\end{table}

\begin{figure*}[t]
	\centering
	\begin{subfigure}{0.3\textwidth}
		\centering
		\includegraphics[width=1\textwidth]{figures/cube3_search_time.eps}
		\caption{RC(12)}
	\end{subfigure}
	\begin{subfigure}{0.3\textwidth}
		\centering
		\includegraphics[width=1\textwidth]{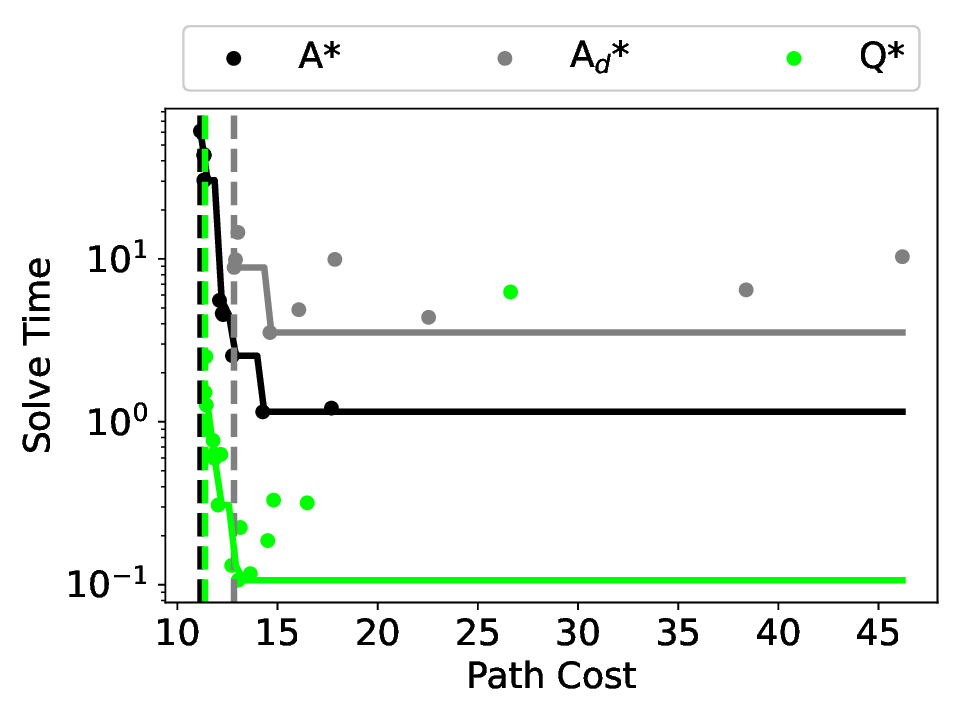}
		\caption{RC(156)}
	\end{subfigure}
	\begin{subfigure}{0.3\textwidth}
		\centering
		\includegraphics[width=1\textwidth]{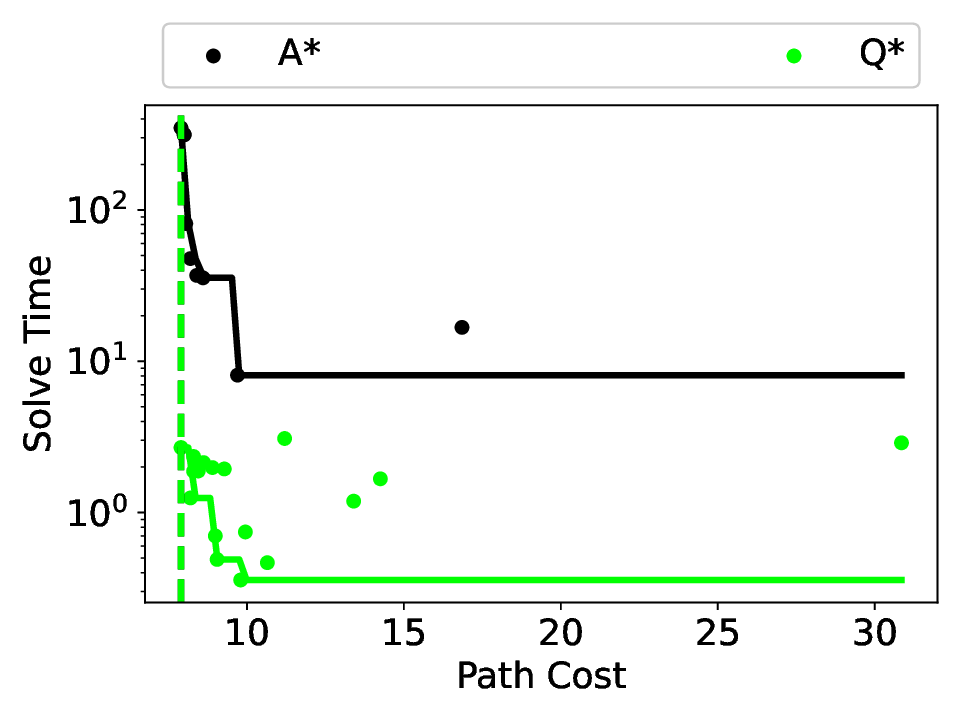}
		\caption{RC(1884)}
	\end{subfigure}
	\caption{Action space size ablation study on Rubik's cube: average path cost vs average time to find a solution.}
	
\label{fig:searchtimeab}
\end{figure*}

\begin{figure*}[t]
\centering
\begin{subfigure}{0.3\textwidth}
	\centering
	\includegraphics[width=1\textwidth]{figures/cube3_search_nodes.eps}
	\caption{RC(12)}
\end{subfigure}
\begin{subfigure}{0.3\textwidth}
	\centering
	\includegraphics[width=1\textwidth]{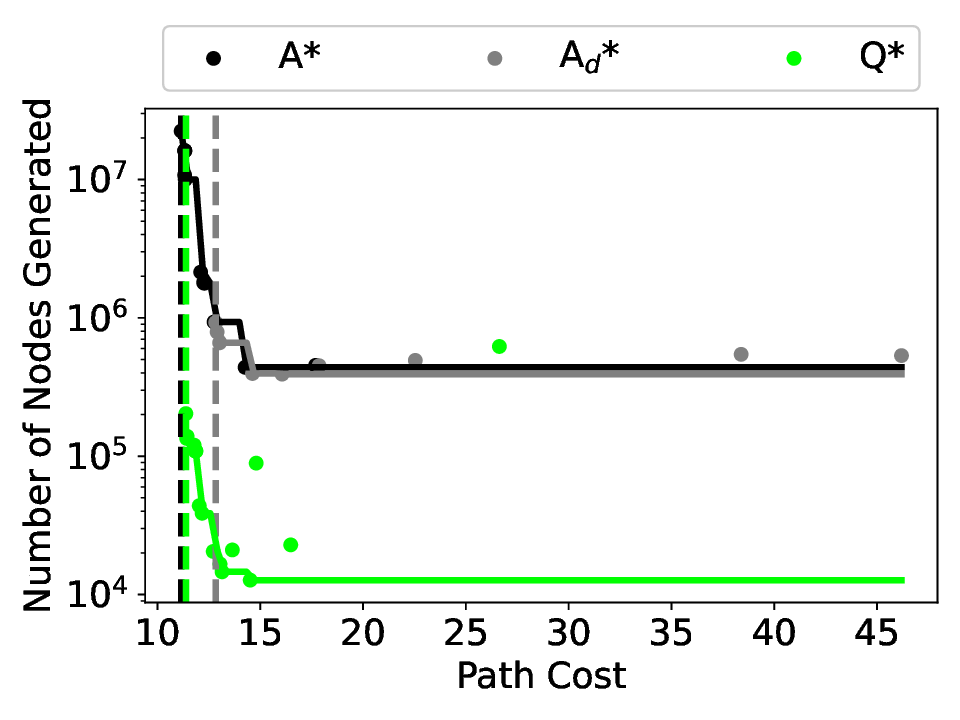}
	\caption{RC(156)}
\end{subfigure}
\begin{subfigure}{0.3\textwidth}
	\centering
	\includegraphics[width=1\textwidth]{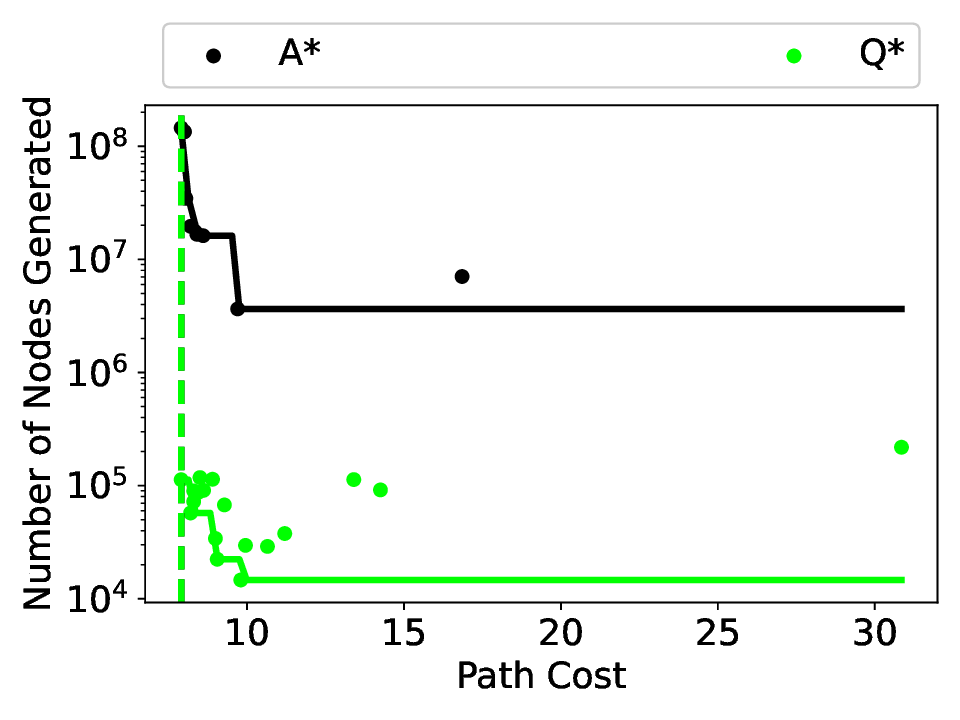}
	\caption{RC(1884)}
\end{subfigure}
\caption{Action space size ablation study on Rubik's cube: average path cost vs average node generations.}
\label{fig:searchnodesab}
\end{figure*}

\begin{table}[tbp]
\centering
\footnotesize
\caption{Performance comparison across different search algorithms and search path cost weights, $\lambda$, averaged over all problem instances and search batch sizes, $b$. The header \% indicates the percentage of problems solved, whereas the header Node counts the average node generation; node counts are reported in thousands (k).}
\label{tab:lambda-analysis}
\setlength{\tabcolsep}{4pt}
\renewcommand{\arraystretch}{1.2}
\begin{tabular}{llrrrrrrrrrr}
	\toprule
	\multirow{2}{*}{Domain} & \multirow{2}{*}{Search} & 
	\multicolumn{2}{c}{$\lambda=0.0$} & \multicolumn{2}{c}{$\lambda=0.2$} & 
	\multicolumn{2}{c}{$\lambda=0.4$} & \multicolumn{2}{c}{$\lambda=0.6$} & 
	\multicolumn{2}{c}{$\lambda=0.8$} \\
	\cmidrule(lr){3-4} \cmidrule(lr){5-6} \cmidrule(lr){7-8} \cmidrule(lr){9-10} \cmidrule(lr){11-12}
	& & \% & Nodes & \% & Nodes & \% & Nodes & \% & Nodes & \% & Nodes \\
	\midrule
	\multirow{3}{*}{RC(12)} 
	& A*     & 100 & \nk{908.667} & 100 & \nk{886.200} & 100 & \nk{1057.333} & \textbf{100} & \nk{5663.333} & 0 & --- \\
	& A$_d$* & 100 & \nk{168.667} & 100 & \nk{99.433}  & 100 & \nk{354.000}  & 67  & \nk{5090.000} & 0 & --- \\
	& Q*     & 100 & \textbf{\nk{127.367}} & 100 & \textbf{\nk{75.170}}  & 100 & \textbf{\nk{75.870}}   & \textbf{100} & \textbf{\nk{213.667}}  & \textbf{100} & \textbf{\nk{3650.000}} \\
	\midrule
	\multirow{3}{*}{35-Pancake} 
	& A*     & 100 & \nk{4122.333} & 100 & \nk{4122.333} & 100 & \nk{4122.333} & 69 & \nk{4089.000} & 92 & \nk{4122.333} \\
	& A$_d$* & 100 & \nk{118.937}  & 100 & \nk{118.937}  & 100 & \nk{118.937}  & \textbf{100} & \nk{118.937}  & \textbf{100} & \nk{118.947} \\
	& Q*     & 100 & \textbf{\nk{118.047}}  & 100 & \nk{118.047}  & 100 & \textbf{\nk{118.047}}  & \textbf{100} & \textbf{\nk{118.047}}  & \textbf{100} & \textbf{\nk{118.047}} \\
	\midrule
	\multirow{3}{*}{LightsOut7} 
	& A*     & 100 & \nk{4086.000} & 100 & \nk{4086.000} & 100 & \nk{4085.667} & 46 & \nk{9030.000} & 0 & --- \\
	& A$_d$* & 100 & \textbf{\nk{85.193}}   & 100 & \textbf{\nk{86.173}}   & 100 & \textbf{\nk{85.027}}   & 34 & \nk{454.000}  & 0 & --- \\
	& Q*     & 100 & \nk{216.000}  & 100 & \nk{106.167}  & 100 & \nk{97.537}   & \textbf{100} & \textbf{\nk{93.247} }  & \textbf{100} & \textbf{\nk{94.943}} \\
	\bottomrule
\end{tabular}
\end{table}

\begin{table}[tbp]
\centering
\footnotesize
\caption{Performance comparison across search algorithms and search batch sizes, $b$, averaged over all problem instances and search path cost weights, $\lambda$. Columns show percentage of problems solved, average node generations (in thousands), and average runtime (seconds). Bold indicates the best value per domain and batch size (except when all values are identical).}
\label{tab:b-analysis}
\setlength{\tabcolsep}{4pt}
\renewcommand{\arraystretch}{1.2}
\begin{tabular}{ll
		r r r
		r r r
		r r r
	}
	\toprule
	\multirow{2}{*}{Domain} & \multirow{2}{*}{Search} & 
	\multicolumn{3}{c}{$b=100$} & \multicolumn{3}{c}{$b=1000$} & \multicolumn{3}{c}{$b=10000$} \\
	\cmidrule(lr){3-5} \cmidrule(lr){6-8} \cmidrule(lr){9-11}
	& & \% & Nodes & Time & \% & Nodes & Time & \% & Nodes & Time \\
	\midrule
	\multirow{3}{*}{RC(12)} 
	& A*     & \textbf{100} & \nk{1414.900} & 6.9 & \textbf{100} & \nk{1576.750} & 6.5 & \textbf{100} & \nk{3395.000} & 12.1 \\
	& A$_d$* &  87 & \nk{1376.950} & 18.9 & \textbf{100} & \nk{1413.375} & 9.1 &  88 & \nk{1493.750} & 8.3 \\
	& Q*     & \textbf{100} & \textbf{\nk{84.855}} & \textbf{1.2} & \textbf{100} & \textbf{\nk{61.700}} & \textbf{0.4} & \textbf{100} & \textbf{\nk{922.000}} & \textbf{6.3} \\
	\midrule
	\multirow{3}{*}{35-Pancake} 
	& A*     & \textbf{100} & \nk{117.000}   & 0.6 & \textbf{100} & \nk{1150.000} & 4.8 &  77 & \nk{11080.000} & 48.7 \\
	& A$_d$* & \textbf{100} & \nk{3.516}     & \textbf{0.1} & \textbf{100} & \nk{33.300}   & \textbf{0.3} & \textbf{100} & \nk{320.000}   & \textbf{2.1} \\
	& Q*     & \textbf{100} & \textbf{\nk{3.340}} & \textbf{0.1} & \textbf{100} & \textbf{\nk{32.800}} & \textbf{0.3} & \textbf{100} & \textbf{\nk{318.000}} & \textbf{2.1} \\
	\midrule
	\multirow{3}{*}{LightsOut7} 
	& A*     & \textbf{100} & \nk{117.667}  & 0.4 &  90 & \nk{2645.000} & 8.6 &  84 & \nk{10975.000} & 29.3 \\
	& A$_d$* & \textbf{100} & \textbf{\nk{4.027}}   & \textbf{0.1} &  84 & \nk{119.025}  & 1.1 &  84 & \nk{315.250}   & 2.0 \\
	& Q*     & \textbf{100} & \nk{62.603}   & 1.2 & \textbf{100} & \textbf{\nk{62.275}}   & \textbf{0.5} & \textbf{100} & \textbf{\nk{274.250}}   & \textbf{1.3} \\
	\bottomrule
\end{tabular}
\end{table}

\begin{table*}[tb]
\centering
\caption{The ratio between A* and \searchalg{} search for the solution time and number of nodes generated for hypothetical acceptable path cost thresholds for RC(12), RC(156), and RC(1884).}
\label{tab:change}
\vskip 0.15in
\begin{tabular}{@{}cccccccccccc@{}}
	\toprule
	& \multicolumn{11}{c}{Path Cost Threshold}                                                                                                         \\ \midrule
	& \multicolumn{3}{c}{\textbf{RC (12)}} & \textbf{} & \multicolumn{3}{c}{\textbf{RC (156)}} & \textbf{} & \multicolumn{3}{c}{\textbf{RC (1884)}} \\ \cmidrule(lr){2-4} \cmidrule(lr){6-8} \cmidrule(l){10-12} 
	& 22         & 25          & 28        &           & 12          & 14         & 16         &           & 8            & 9          & 10         \\ \cmidrule(lr){2-4} \cmidrule(lr){6-8} \cmidrule(l){10-12} 
	Time  & 0.8        & 5.1         & 1.7       &           & 50.8        & 23.9       & 10.8       &           & 129.7        & 28.5       & 22.6       \\
	Nodes & 1.4        & 11.9        & 6.8       &           & 92.0        & 64.0       & 34.5       &           & 1288.4       & 282.8      & 249.1      \\ \bottomrule
\end{tabular}
\end{table*}

\begin{table}[tb]
\centering
\caption{Ratio of action space sizes, along with performance ratios (time and nodes generated) compared to RC(12), averaged over all search parameter settings, with standard deviation in parenthesis.}
\label{tab:relperf}
\vskip 0.15in
\begin{tabular}{@{}ccccc@{}}
	\toprule
	Puzzle                    & Actions               & Method & Time              & Nodes Gen         \\ \midrule
	\multirow{2}{*}{RC(156)}  & \multirow{2}{*}{x13}  & A*    & 3.5(1.6)          & 8.7(2.2)          \\
	&                       & \searchalg{}    & \textbf{0.9(0.7)} & \textbf{1.4(1.3)} \\ \midrule
	\multirow{2}{*}{RC(1884)} & \multirow{2}{*}{x157} & A*    & 37.0(6.5)         & 62.7(5.2)         \\
	&                       & \searchalg{}    & \textbf{3.7(4.0)} & \textbf{2.3(3.6)} \\ \bottomrule
\end{tabular}
\end{table}

The results are reported in Figures \ref{fig:searchtime} and \ref{fig:searchnodes}.  These figures illustrate the relationship between the average path cost and either the average time taken to find a solution or the average number of generated nodes, respectively. Both figures employ a logarithmic scale on the y-axis, with each data point representing a specific search parameter setting. The dashed line signifies the lowest average path cost identified, while the solid line represents either the fastest solution time or the fewest number of node generations, depending on the context, as determined by a hypothetical threshold for an acceptable average path cost. Detailed results from each search run are shown in the Appendix.

The figures show that, for almost any possible path cost threshold, \searchalg{} is significantly faster and generates significantly fewer nodes than both A* and A$_d$*. A$_d$* exhibits overall improvement over A* as it generates only one node per iteration. However, as it assigns the same $f$-value to all children of a node, it introduces inefficiencies by being unable to prioritize one child node over another.

The lowest average path cost achieved by all algorithms remains comparable, with the maximum difference in average path cost being a mere $0.1\%$, observed in the RC domain. Overall, for RC, in the best case, A* finds a shortest path 59\% of the time while \searchalg{} finds a shortest path 56.4\% of the time. For Lights Out, in the best case, both A* and \searchalg{} find a shortest path 100\% of the time.

We also include an analysis of the results as a function of the weight on the path cost $\lambda$ and the search batch size $b$ across the different domains, among all configurations for which all algorithms solved at least one problem. Table~\ref{tab:lambda-analysis} reports the percentage of problems solved and the average number of node generations across all problem instances and batch sizes for each tested value of $\lambda$. In the Rubik's Cube domain, \searchalg{} consistently outperforms both A* and A$_d$* across all configurations, expanding up to 25$\times$ fewer nodes at $\lambda = 0.6$, and is the only approach capable of solving problems under the strictest optimality requirement of $\lambda = 0.8$. In 35-Pancake, \searchalg{} and A$_d$* perform similarly, with a slight advantage for \searchalg{}; both substantially outperform standard A*. Lastly, in LightsOut, A$_d$* performs better than \searchalg{} for lower values of $\lambda$ (0--0.4), but its performance deteriorates significantly at higher values: at $\lambda = 0.6$, it expands over 4.5$\times$ more nodes than \searchalg{}, and at $\lambda = 0.8$, \searchalg{} is the only method able to solve all problem instances, while the others fail to solve any.

Table~\ref{tab:b-analysis} presents results averaged across all problem instances for each tested batch size \( b \). The trends observed earlier remain consistent across different values of \( b \). In general, the advantage of \searchalg{} becomes more pronounced as \( b \) increases. While smaller values of \( b \) often lead to fewer node expansions overall, they do not always yield the lowest runtime. For example, in the Rubik’s Cube domain, the smallest runtime is achieved at \( b = 1000 \), whereas in the Pancake puzzle, \( b = 100 \) yields the best performance.

\subsection{Ablation Study: Varying Number of Actions}

\begin{figure*}[t]
\centering
\begin{subfigure}{0.3\textwidth}
	\centering
	\includegraphics[width=1\textwidth]{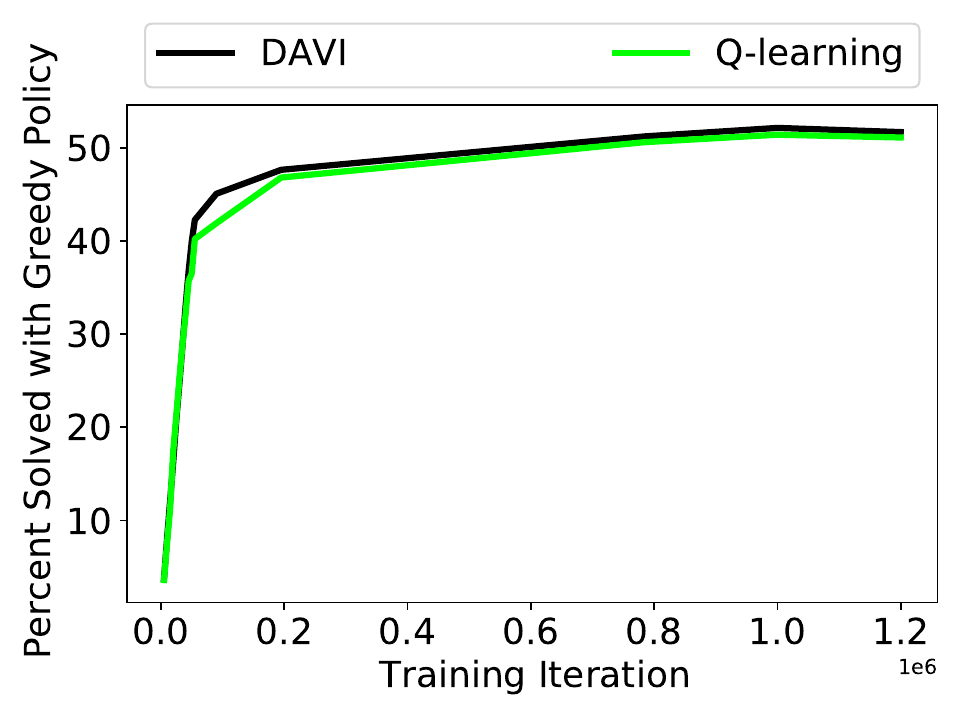}
	\caption{RC(12)}
\end{subfigure}
~
\begin{subfigure}{0.3\textwidth}
	\centering
	\includegraphics[width=1\textwidth]{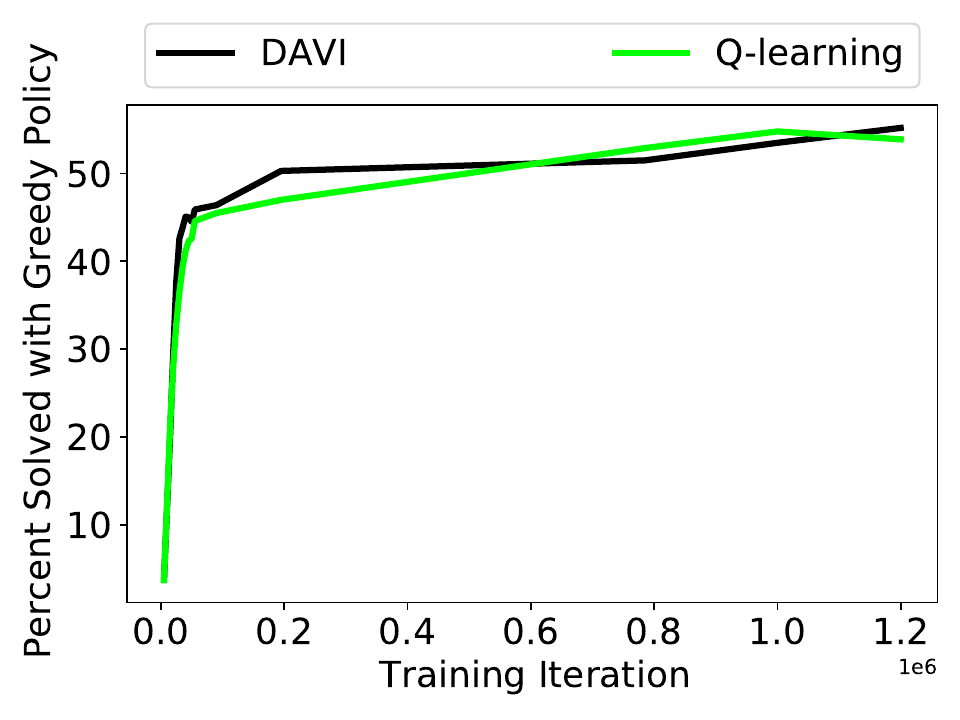}
	\caption{RC(156)}
\end{subfigure}
~
\begin{subfigure}{0.3\textwidth}
	\centering
	\includegraphics[width=1\textwidth]{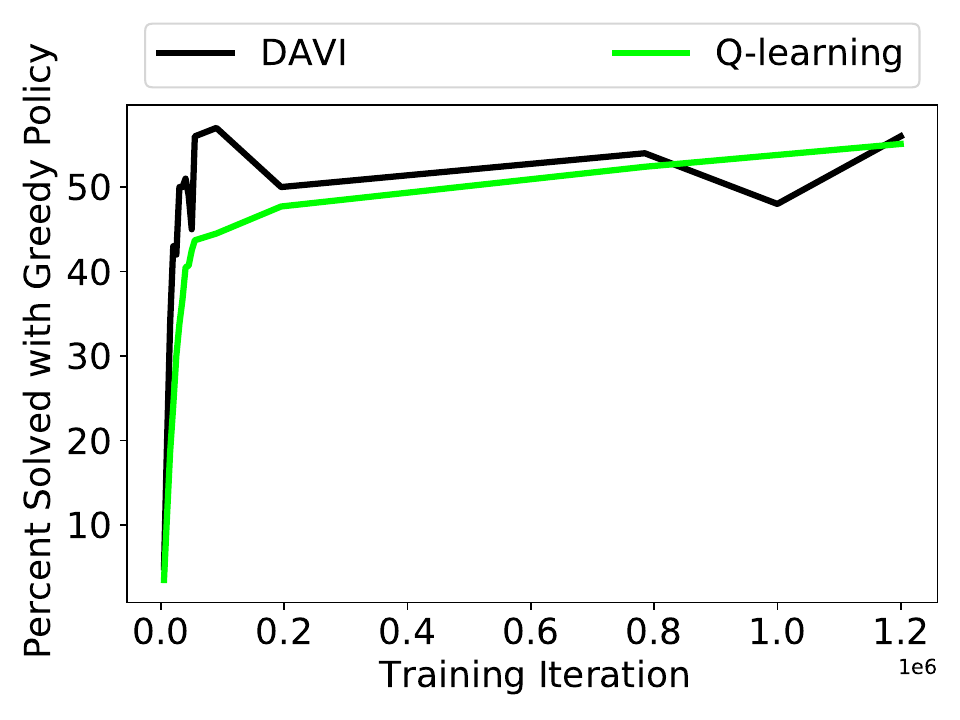}
	\caption{RC(1884)}
\end{subfigure}
\caption{Percentage of training states solved with a greedy policy as a function of training iteration.}
\label{fig:train}
\end{figure*}

In order to study the performance of \searchalg{} as the number of actions increases, we perform an ablation study focusing on the Rubik's cube domain.
The standard RC action space includes 12 different actions: each of the six faces can be turned clockwise or counterclockwise. We denote this action space as RC(12). For the ablation study, we add meta-actions to the RC action space, creating RC(156) and RC(1884). RC(156) has all the actions in RC(12) plus all combinations of actions of size two, RC(144). RC(1884) has all the actions in RC(156) plus all combinations of actions of size three (1728). To ensure none of these additional meta-actions are redundant, the cost for all meta-actions is also set to one.

In the experiments reported earlier, we trained the model for 1.2 million iterations.
However, as the size of the action space increases, training becomes infeasible for DAVI. Table \ref{tab:traintime} shows that DAVI would take over a month to train on RC(156) and almost a year to train on RC(1884). Therefore, we reduce the training batch size ($TB$) in proportion to the differences in the size of the action space with RC(12). Since RC(156) has 13 times more actions, we train DAVI with a $TB=769$, and since RC(1884) has 157 times more actions, we train DAVI with a training batch size of $TB=63$. Moreover given the escalation in solving time with the expansion of the action space in A* search, we utilize a subset of 100 states for RC(156) and 20 states for RC(1884), instead of the full 500 states employed for RC(12).

Figures~\ref{fig:searchtimeab} and \ref{fig:searchnodesab} repeat the experiments of Figures~\ref{fig:searchtime} and \ref{fig:searchnodes} on the RC environment with the different action-space sizes. The results show that the performance of \searchalg{} over A* and A*$_d$ becomes even more pronounced as the action space increases. In fact, for RC(1884), A$_d$* was unable to find a solution due to running out of memory.
In the most extreme case, the cheapest average path cost for A* and \searchalg{} is identical for RC(1884), however, \searchalg{} is 129 times faster and generates 1228 times fewer nodes than A*. The ratios for various desired average path costs are presented in Table \ref{tab:change}. It is evident from the table that \searchalg{} consistently outperforms A* in all instances except one, often exhibiting orders of magnitude faster speed and greater memory efficiency than A*.



When comparing A* and \searchalg{} to themselves for different action spaces, Table \ref{tab:relperf} shows that, though RC(1884) has 157 times more actions than RC(12), \searchalg{} only takes 3.7 times as long to find a solution and generates only 2.3 times as many nodes. On the other hand, in this same scenario, A* takes 37 times as long and generates 62.7 times as many nodes.
Overall, \searchalg{} has much better performance for both metrics. For RC(156) \searchalg{} finds solutions in even less time than it did for RC(12) due to the addition of meta-actions.

\subsubsection{Performance During Training}
To monitor performance during training, we track the percentage of states that are solved by simply behaving greedily with respect to the cost-to-go function. We generate these states the same way we generate the training states. Figure \ref{fig:train} shows this metric as a function of training time. The results show that, in RC(12), DAVI is slightly better than Q-learning. In RC(156) and RC(1884), even though the training batch size for DAVI is smaller, the performance is on par with Q-learning. This may be due to the fact that DAVI is only learning the cost-to-go for a single state while Q-learning must learn the sum of the transition cost and cost-to-go for all possible next states.

\section{Discussion}
\label{sec:discuss}
As the size of the action space increases, \searchalg{} becomes significantly more effective than A* in terms of solution time and the number of nodes generated. In the largest action space \searchalg{} is orders of magnitude faster and generates orders of magnitude fewer nodes than A* while finding solutions with the same average path cost. For smaller action spaces, while \searchalg{} is almost always faster and more memory efficient, A* is capable of finding solutions that are slightly cheaper than \searchalg{}. This could be due to the difference in what $\ctgfunc$ and $\dqnfunc$ are computing. Since the forward pass performed by the DQN, $\dqnfunc$, is the same as doing a one-step lookahead with $\ctgfunc$, this could make learning $\dqnfunc$ more difficult than learning $\ctgfunc$. This may explain why, in the case of Lights Out, A$_d$* is, in some cases, faster and generates fewer nodes than \searchalg{}. However, \searchalg{} becomes better as the path cost threshold decreases. Since training and search are significantly faster for Q-learning and \searchalg{}, this gap could be closed with longer training times and searching with larger values of $\lambda$ or $N$.

While the DQN used in our work was designed for environments with a fixed set of actions, the \searchalg{} algorithm itself is not limited in this way. It can naturally be applied to environments with a dynamic action space---where the set of available actions changes from one state to another---provided that a DQN (or similar function approximator) is available that can compute Q-values for the actions available in the current state.
Constructing such a DQN requires the ability to evaluate actions that are not drawn from a fixed, static list. One effective strategy is to exploit domain structure to represent actions in a way that generalizes across states. For example, Graph Convolutional Policy Networks (GCPNs) \cite{you2018graph} were developed to generate molecular graphs by sequentially adding atoms and bonds. In this setting, the action space is dynamic, since each molecule's current structure determines which atoms and bonds can legally be added next. GCPNs overcome this by representing the molecular structure as a graph and using Graph Convolutional Networks (GCNs) to compute node embeddings that reflect the local and global structure. These embeddings are then used to predict valid actions, such as which atom to add or where to place a bond.
A similar idea can be applied to Q-learning: instead of representing Q-values for a fixed number of actions for all states, a graph-based neural network can be trained to compute Q-values for the dynamically determined set of available actions, using representations derived from the current state and action candidates. This allows \searchalg{} to be used in environments with complex, structured, and dynamically evolving action spaces.
In problems involving sequences, Long Short-Term Memory \cite{hochreiter1997long} or Transformer \cite{vaswani2017attention} architectures could be used to compute Q-factors. This would have a direct application to problems with large, but dynamic, action spaces such as chemical synthesis, theorem proving, program synthesis, and web navigation.


\section{Conclusion}
Efficiently solving search problems with large action spaces has been of importance to the artificial intelligence community for decades \cite{russell1992efficient, korf1993linear, yoshizumi2000partial}. \searchalg{} search uses state-action based heuristic estimations to eliminate the majority of the computational and memory burden associated with large action spaces by generating only one node per iteration and requiring only one application of the heuristic function per iteration. To obtain such state-action estimates, we trained a deep Q-network (DQN) using a training scheme that adapts Q-learning for deterministic finite-horizon environments.
When compared to A* search, \searchalg{} search, equipped with the trained DQN, is up to 129 times faster and generates up to 1288 times fewer nodes. When increasing the size of the action space by 157 times, \searchalg{} search only takes 3.7 times as long and generates only 2.3 times more nodes. The ability that \searchalg{} has to efficiently scale up to large action spaces could play a significant role in finding solutions to many important problems with large action spaces.

\section*{Acknowledgements}
The work of Pierre Baldi is in part supported by NIH grant: R01GM123558.
The work of Shahaf Shperberg was supported by the Israel Science Foundation (ISF)
grant \#909/23,  by Israel's Ministry of Innovation, Science and Technology (MOST) grant \#1001706842, in collaboration with Israel National Road Safety Authority and Netivei Israel, awarded to Shahaf Shperberg, by BSF grant \#2024614 awarded to Shahaf Shperberg, as part of a joint NSF-BSF grant with Forest Agostinelli. This material is based upon work supported by the National Science Foundation under Award No. 2426622.

\bibliographystyle{elsarticle-num-names} 
\bibliography{mybib}

\pagebreak
\appendix

\section*{Appendix}

Results for all search parameter settings for the Rubik's cube, LightsOut7, and the 35-Pancake puzzle are shown in Tables \ref{tab:rcresultsfull}, \ref{tab:lightsout7resultsfull}, and \ref{tab:35pancakeresultsfull}, respectively along the dimensions of average path cost, percentage of problems solved, average number of nodes generated, average time taken to find a solution, and average nodes per second. All averages are only computed for solved states.

\begin{longtable}{|l|l|l|l|l|l|l|}
\hline
Params       & Search & Path Cost & \% Solved & Nodes    & Secs  & Nodes/Sec \\ \hline
w1.0\_b100   & A*     & -         & -         & -        & -     & -         \\ \hline
w1.0\_b100   & A$_d$* & -         & -         & -        & -     & -         \\ \hline
w1.0\_b100   & Q*     & -         & -         & -        & -     & -         \\ \hline
w1.0\_b1000  & A*     & -         & -         & -        & -     & -         \\ \hline
w1.0\_b1000  & A$_d$* & -         & -         & -        & -     & -         \\ \hline
w1.0\_b1000  & Q*     & -         & -         & -        & -     & -         \\ \hline
w1.0\_b10000 & A*     & -         & -         & -        & -     & -         \\ \hline
w1.0\_b10000 & A$_d$* & -         & -         & -        & -     & -         \\ \hline
w1.0\_b10000 & Q*     & -         & -         & -        & -     & -         \\ \hline
w0.8\_b100   & A*     & -         & -         & -        & -     & -         \\ \hline
w0.8\_b100   & A$_d$* & -         & -         & -        & -     & -         \\ \hline
w0.8\_b100   & Q*     & 21.53     & 100       & 3.61E+06 & 58.41 & 6.59E+04  \\ \hline
w0.8\_b1000  & A*     & -         & -         & -        & -     & -         \\ \hline
w0.8\_b1000  & A$_d$* & -         & -         & -        & -     & -         \\ \hline
w0.8\_b1000  & Q*     & 21.53     & 100       & 3.62E+06 & 28.74 & 1.48E+05  \\ \hline
w0.8\_b10000 & A*     & -         & -         & -        & -     & -         \\ \hline
w0.8\_b10000 & A$_d$* & -         & -         & -        & -     & -         \\ \hline
w0.8\_b10000 & Q*     & 21.53     & 100       & 3.72E+06 & 27.47 & 1.60E+05  \\ \hline
w0.6\_b100   & A*     & 21.5      & 100       & 5.17E+06 & 25.13 & 2.12E+05  \\ \hline
w0.6\_b100   & A$_d$* & 21.47     & 49.1      & 5.00E+06 & 68.75 & 7.59E+04  \\ \hline
w0.6\_b100   & Q*     & 22.61     & 100       & 1.64E+05 & 2.25  & 7.13E+04  \\ \hline
w0.6\_b1000  & A*     & 21.5      & 100       & 5.29E+06 & 22.2  & 2.50E+05  \\ \hline
w0.6\_b1000  & A$_d$* & 21.5      & 100       & 5.17E+06 & 33.64 & 1.65E+05  \\ \hline
w0.6\_b1000  & Q*     & 22.61     & 100       & 1.76E+05 & 1.12  & 1.61E+05  \\ \hline
w0.6\_b10000 & A*     & 21.49     & 100       & 6.53E+06 & 24.61 & 2.69E+05  \\ \hline
w0.6\_b10000 & A$_d$* & 21.48     & 51.1      & 5.10E+06 & 29    & 1.87E+05  \\ \hline
w0.6\_b10000 & Q*     & 22.52     & 100       & 3.01E+05 & 1.57  & 1.98E+05  \\ \hline
w0.4\_b100   & A*     & 23.5      & 100       & 3.21E+05 & 1.44  & 2.17E+05  \\ \hline
w0.4\_b100   & A$_d$* & 23.59     & 100       & 3.05E+05 & 3.89  & 7.80E+04  \\ \hline
w0.4\_b100   & Q*     & 25.03     & 100       & 7.41E+03 & 0.21  & 3.89E+04  \\ \hline
w0.4\_b1000  & A*     & 23.21     & 100       & 4.81E+05 & 1.74  & 2.80E+05  \\ \hline
w0.4\_b1000  & A$_d$* & 23.52     & 100       & 3.17E+05 & 1.8   & 1.80E+05  \\ \hline
w0.4\_b1000  & Q*     & 24.13     & 100       & 2.32E+04 & 0.19  & 1.37E+05  \\ \hline
w0.4\_b10000 & A*     & 22.27     & 100       & 2.37E+06 & 8.09  & 2.93E+05  \\ \hline
w0.4\_b10000 & A$_d$* & 23.27     & 100       & 4.40E+05 & 2.09  & 2.15E+05  \\ \hline
w0.4\_b10000 & Q*     & 22.49     & 100       & 1.97E+05 & 0.82  & 2.40E+05  \\ \hline
w0.2\_b100   & A*     & 27.28     & 100       & 5.06E+04 & 0.31  & 1.65E+05  \\ \hline
w0.2\_b100   & A$_d$* & 29.3      & 100       & 3.58E+04 & 0.57  & 5.72E+04  \\ \hline
w0.2\_b100   & Q*     & 28.18     & 100       & 7.01E+03 & 0.21  & 3.25E+04  \\ \hline
w0.2\_b1000  & A*     & 24.13     & 100       & 2.68E+05 & 0.95  & 2.82E+05  \\ \hline
w0.2\_b1000  & A$_d$* & 27.48     & 100       & 4.55E+04 & 0.33  & 1.38E+05  \\ \hline
w0.2\_b1000  & Q*     & 24.2      & 100       & 2.25E+04 & 0.18  & 1.36E+05  \\ \hline
w0.2\_b10000 & A*     & 22.35     & 100       & 2.34E+06 & 7.83  & 2.98E+05  \\ \hline
w0.2\_b10000 & A$_d$* & 24.39     & 100       & 2.17E+05 & 0.97  & 2.23E+05  \\ \hline
w0.2\_b10000 & Q*     & 22.49     & 100       & 1.96E+05 & 0.79  & 2.48E+05  \\ \hline
w0.0\_b100   & A*     & 38.06     & 100       & 1.18E+05 & 0.61  & 1.57E+05  \\ \hline
w0.0\_b100   & A$_d$* & 63.79     & 100       & 1.67E+05 & 2.42  & 4.63E+04  \\ \hline
w0.0\_b100   & Q*     & 38.95     & 100       & 1.61E+05 & 2.12  & 4.57E+04  \\ \hline
w0.0\_b1000  & A*     & 24.19     & 100       & 2.68E+05 & 0.95  & 2.84E+05  \\ \hline
w0.0\_b1000  & A$_d$* & 39.78     & 100       & 1.21E+05 & 0.76  & 1.35E+05  \\ \hline
w0.0\_b1000  & Q*     & 24.41     & 100       & 2.51E+04 & 0.2   & 1.38E+05  \\ \hline
w0.0\_b10000 & A*     & 22.35     & 100       & 2.34E+06 & 7.83  & 2.98E+05  \\ \hline
w0.0\_b10000 & A$_d$* & 24.57     & 100       & 2.18E+05 & 0.98  & 2.22E+05  \\ \hline
w0.0\_b10000 & Q*     & 22.49     & 100       & 1.96E+05 & 0.79  & 2.49E+05  \\ \hline
\caption{Results for all search parameter settings for the Rubik's cube.}
\label{tab:rcresultsfull}
\end{longtable}

\begin{longtable}{|l|l|l|l|l|l|l|}
\hline
Params       & Search & Path Cost & \% Solved & Nodes    & Secs  & Nodes/Sec \\ \hline
w1.0\_b100   & A*     & -         & -         & -        & -     & -         \\ \hline
w1.0\_b100   & A$_d$* & -         & -         & -        & -     & -         \\ \hline
w1.0\_b100   & Q*     & -         & -         & -        & -     & -         \\ \hline
w1.0\_b1000  & A*     & -         & -         & -        & -     & -         \\ \hline
w1.0\_b1000  & A$_d$* & -         & -         & -        & -     & -         \\ \hline
w1.0\_b1000  & Q*     & -         & -         & -        & -     & -         \\ \hline
w1.0\_b10000 & A*     & -         & -         & -        & -     & -         \\ \hline
w1.0\_b10000 & A$_d$* & -         & -         & -        & -     & -         \\ \hline
w1.0\_b10000 & Q*     & -         & -         & -        & -     & -         \\ \hline
w0.8\_b100   & A*     & -         & -         & -        & -     & -         \\ \hline
w0.8\_b100   & A$_d$* & -         & -         & -        & -     & -         \\ \hline
w0.8\_b100   & Q*     & 24.26     & 100       & 6.63E+03 & 0.16  & 3.68E+04  \\ \hline
w0.8\_b1000  & A*     & -         & -         & -        & -     & -         \\ \hline
w0.8\_b1000  & A$_d$* & -         & -         & -        & -     & -         \\ \hline
w0.8\_b1000  & Q*     & 24.26     & 100       & 3.42E+04 & 0.26  & 1.53E+05  \\ \hline
w0.8\_b10000 & A*     & -         & -         & -        & -     & -         \\ \hline
w0.8\_b10000 & A$_d$* & -         & -         & -        & -     & -         \\ \hline
w0.8\_b10000 & Q*     & 24.26     & 100       & 2.44E+05 & 1.08  & 2.48E+05  \\ \hline
w0.6\_b100   & A*     & -         & -         & -        & -     & -         \\ \hline
w0.6\_b100   & A$_d$* & 23.84     & 34.2      & 3.82E+05 & 8.13  & 5.14E+04  \\ \hline
w0.6\_b100   & Q*     & 24.28     & 100       & 4.94E+03 & 0.12  & 3.72E+04  \\ \hline
w0.6\_b1000  & A*     & 24.16     & 57.8      & 7.16E+06 & 24.62 & 3.42E+05  \\ \hline
w0.6\_b1000  & A$_d$* & 23.84     & 34.2      & 3.99E+05 & 3.82  & 1.38E+05  \\ \hline
w0.6\_b1000  & Q*     & 24.29     & 100       & 3.38E+04 & 0.25  & 1.55E+05  \\ \hline
w0.6\_b10000 & A*     & 23.84     & 34.2      & 1.09E+07 & 29.52 & 3.72E+05  \\ \hline
w0.6\_b10000 & A$_d$* & 23.84     & 34.2      & 5.81E+05 & 3.94  & 1.69E+05  \\ \hline
w0.6\_b10000 & Q*     & 24.27     & 100       & 2.41E+05 & 1.05  & 2.48E+05  \\ \hline
w0.4\_b100   & A*     & 24.26     & 100       & 1.17E+05 & 0.42  & 2.82E+05  \\ \hline
w0.4\_b100   & A$_d$* & 24.33     & 100       & 4.18E+03 & 0.11  & 3.54E+04  \\ \hline
w0.4\_b100   & Q*     & 24.31     & 100       & 9.71E+03 & 0.21  & 3.65E+04  \\ \hline
w0.4\_b1000  & A*     & 24.26     & 100       & 1.14E+06 & 3.22  & 3.56E+05  \\ \hline
w0.4\_b1000  & A$_d$* & 24.27     & 100       & 2.49E+04 & 0.21  & 1.22E+05  \\ \hline
w0.4\_b1000  & Q*     & 24.34     & 100       & 3.79E+04 & 0.3   & 1.53E+05  \\ \hline
w0.4\_b10000 & A*     & 24.26     & 100       & 1.10E+07 & 29.11 & 3.79E+05  \\ \hline
w0.4\_b10000 & A$_d$* & 24.26     & 100       & 2.26E+05 & 1.3   & 1.74E+05  \\ \hline
w0.4\_b10000 & Q*     & 24.3      & 100       & 2.45E+05 & 1.09  & 2.47E+05  \\ \hline
w0.2\_b100   & A*     & 24.27     & 100       & 1.18E+05 & 0.42  & 2.82E+05  \\ \hline
w0.2\_b100   & A$_d$* & 24.5      & 100       & 5.12E+03 & 0.13  & 3.49E+04  \\ \hline
w0.2\_b100   & Q*     & 24.38     & 100       & 1.51E+04 & 0.31  & 3.73E+04  \\ \hline
w0.2\_b1000  & A*     & 24.26     & 100       & 1.14E+06 & 3.2   & 3.59E+05  \\ \hline
w0.2\_b1000  & A$_d$* & 24.3      & 100       & 2.64E+04 & 0.22  & 1.23E+05  \\ \hline
w0.2\_b1000  & Q*     & 24.42     & 100       & 4.64E+04 & 0.42  & 1.50E+05  \\ \hline
w0.2\_b10000 & A*     & 24.26     & 100       & 1.10E+07 & 29.13 & 3.79E+05  \\ \hline
w0.2\_b10000 & A$_d$* & 24.27     & 100       & 2.27E+05 & 1.33  & 1.72E+05  \\ \hline
w0.2\_b10000 & Q*     & 24.38     & 100       & 2.57E+05 & 1.16  & 2.48E+05  \\ \hline
w0.0\_b100   & A*     & 24.32     & 100       & 1.18E+05 & 0.43  & 2.79E+05  \\ \hline
w0.0\_b100   & A$_d$* & 24.64     & 100       & 2.78E+03 & 0.09  & 3.40E+04  \\ \hline
w0.0\_b100   & Q*     & 25.12     & 100       & 1.63E+05 & 2.96  & 3.39E+04  \\ \hline
w0.0\_b1000  & A*     & 24.26     & 100       & 1.14E+06 & 3.22  & 3.56E+05  \\ \hline
w0.0\_b1000  & A$_d$* & 24.37     & 100       & 2.58E+04 & 0.22  & 1.22E+05  \\ \hline
w0.0\_b1000  & Q*     & 25.22     & 100       & 1.31E+05 & 1.16  & 1.48E+05  \\ \hline
w0.0\_b10000 & A*     & 24.26     & 100       & 1.10E+07 & 29.28 & 3.76E+05  \\ \hline
w0.0\_b10000 & A$_d$* & 24.32     & 100       & 2.27E+05 & 1.33  & 1.71E+05  \\ \hline
w0.0\_b10000 & Q*     & 25.59     & 100       & 3.54E+05 & 1.9   & 2.38E+05  \\ \hline
\caption{Results for all search parameter settings for LightsOut7.}
\label{tab:lightsout7resultsfull}
\end{longtable}

\begin{longtable}{|l|l|l|l|l|l|l|}
\hline
Params       & Search & Path Cost & \% Solved & Nodes    & Secs  & Nodes/Sec \\ \hline
w1.0\_b100   & A*     & -         & -         & -        & -     & -         \\ \hline
w1.0\_b100   & A$_d$* & -         & -         & -        & -     & -         \\ \hline
w1.0\_b100   & Q*     & -         & -         & -        & -     & -         \\ \hline
w1.0\_b1000  & A*     & -         & -         & -        & -     & -         \\ \hline
w1.0\_b1000  & A$_d$* & -         & -         & -        & -     & -         \\ \hline
w1.0\_b1000  & Q*     & -         & -         & -        & -     & -         \\ \hline
w1.0\_b10000 & A*     & -         & -         & -        & -     & -         \\ \hline
w1.0\_b10000 & A$_d$* & -         & -         & -        & -     & -         \\ \hline
w1.0\_b10000 & Q*     & -         & -         & -        & -     & -         \\ \hline
w0.8\_b100   & A*     & 33.95     & 100       & 1.17E+05 & 0.56  & 2.09E+05  \\ \hline
w0.8\_b100   & A$_d$* & 35.65     & 100       & 3.54E+03 & 0.12  & 3.67E+04  \\ \hline
w0.8\_b100   & Q*     & 33.99     & 100       & 3.34E+03 & 0.12  & 3.44E+04  \\ \hline
w0.8\_b1000  & A*     & 33.76     & 100       & 1.15E+06 & 4.82  & 2.38E+05  \\ \hline
w0.8\_b1000  & A$_d$* & 34.24     & 100       & 3.33E+04 & 0.28  & 1.22E+05  \\ \hline
w0.8\_b1000  & Q*     & 33.76     & 100       & 3.28E+04 & 0.29  & 1.15E+05  \\ \hline
w0.8\_b10000 & A*     & 33.64     & 76.6      & 1.11E+07 & 48.17 & 2.31E+05  \\ \hline
w0.8\_b10000 & A$_d$* & 33.83     & 100       & 3.20E+05 & 2.08  & 1.54E+05  \\ \hline
w0.8\_b10000 & Q*     & 33.7      & 100       & 3.18E+05 & 2.22  & 1.44E+05  \\ \hline
w0.6\_b100   & A*     & 33.95     & 100       & 1.17E+05 & 0.56  & 2.10E+05  \\ \hline
w0.6\_b100   & A$_d$* & 35.75     & 100       & 3.51E+03 & 0.11  & 3.62E+04  \\ \hline
w0.6\_b100   & Q*     & 33.99     & 100       & 3.34E+03 & 0.11  & 3.51E+04  \\ \hline
w0.6\_b1000  & A*     & 33.76     & 100       & 1.15E+06 & 4.84  & 2.37E+05  \\ \hline
w0.6\_b1000  & A$_d$* & 34.24     & 100       & 3.33E+04 & 0.28  & 1.22E+05  \\ \hline
w0.6\_b1000  & Q*     & 33.76     & 100       & 3.28E+04 & 0.29  & 1.16E+05  \\ \hline
w0.6\_b10000 & A*     & 33.37     & 7.6       & 1.10E+07 & 49.23 & 2.24E+05  \\ \hline
w0.6\_b10000 & A$_d$* & 33.84     & 100       & 3.20E+05 & 2.1   & 1.52E+05  \\ \hline
w0.6\_b10000 & Q*     & 33.7      & 100       & 3.18E+05 & 2.21  & 1.44E+05  \\ \hline
w0.4\_b100   & A*     & 33.95     & 100       & 1.17E+05 & 0.56  & 2.10E+05  \\ \hline
w0.4\_b100   & A$_d$* & 35.75     & 100       & 3.51E+03 & 0.11  & 3.72E+04  \\ \hline
w0.4\_b100   & Q*     & 33.99     & 100       & 3.34E+03 & 0.11  & 3.52E+04  \\ \hline
w0.4\_b1000  & A*     & 33.76     & 100       & 1.15E+06 & 4.84  & 2.37E+05  \\ \hline
w0.4\_b1000  & A$_d$* & 34.24     & 100       & 3.33E+04 & 0.28  & 1.21E+05  \\ \hline
w0.4\_b1000  & Q*     & 33.76     & 100       & 3.28E+04 & 0.29  & 1.15E+05  \\ \hline
w0.4\_b10000 & A*     & 33.69     & 100       & 1.11E+07 & 48.57 & 2.29E+05  \\ \hline
w0.4\_b10000 & A$_d$* & 33.84     & 100       & 3.20E+05 & 2.08  & 1.54E+05  \\ \hline
w0.4\_b10000 & Q*     & 33.7      & 100       & 3.18E+05 & 2.24  & 1.42E+05  \\ \hline
w0.2\_b100   & A*     & 33.95     & 100       & 1.17E+05 & 0.56  & 2.10E+05  \\ \hline
w0.2\_b100   & A$_d$* & 35.75     & 100       & 3.51E+03 & 0.11  & 3.63E+04  \\ \hline
w0.2\_b100   & Q*     & 33.99     & 100       & 3.34E+03 & 0.11  & 3.49E+04  \\ \hline
w0.2\_b1000  & A*     & 33.76     & 100       & 1.15E+06 & 4.85  & 2.37E+05  \\ \hline
w0.2\_b1000  & A$_d$* & 34.24     & 100       & 3.33E+04 & 0.28  & 1.21E+05  \\ \hline
w0.2\_b1000  & Q*     & 33.76     & 100       & 3.28E+04 & 0.29  & 1.14E+05  \\ \hline
w0.2\_b10000 & A*     & 33.69     & 100       & 1.11E+07 & 48.68 & 2.29E+05  \\ \hline
w0.2\_b10000 & A$_d$* & 33.84     & 100       & 3.20E+05 & 2.1   & 1.52E+05  \\ \hline
w0.2\_b10000 & Q*     & 33.7      & 100       & 3.18E+05 & 2.24  & 1.42E+05  \\ \hline
w0.0\_b100   & A*     & 33.95     & 100       & 1.17E+05 & 0.56  & 2.10E+05  \\ \hline
w0.0\_b100   & A$_d$* & 35.75     & 100       & 3.51E+03 & 0.12  & 3.64E+04  \\ \hline
w0.0\_b100   & Q*     & 33.99     & 100       & 3.34E+03 & 0.12  & 3.44E+04  \\ \hline
w0.0\_b1000  & A*     & 33.76     & 100       & 1.15E+06 & 4.85  & 2.37E+05  \\ \hline
w0.0\_b1000  & A$_d$* & 34.24     & 100       & 3.33E+04 & 0.28  & 1.22E+05  \\ \hline
w0.0\_b1000  & Q*     & 33.76     & 100       & 3.28E+04 & 0.29  & 1.15E+05  \\ \hline
w0.0\_b10000 & A*     & 33.69     & 100       & 1.11E+07 & 48.71 & 2.29E+05  \\ \hline
w0.0\_b10000 & A$_d$* & 33.84     & 100       & 3.20E+05 & 2.11  & 1.52E+05  \\ \hline
w0.0\_b10000 & Q*     & 33.7      & 100       & 3.18E+05 & 2.21  & 1.44E+05  \\ \hline
\caption{Results for all search parameter settings for the 35-Pancake puzzle.}
\label{tab:35pancakeresultsfull}
\end{longtable}

\end{document}